%% file: main.tex
\documentclass[sigconf]{aamas} 

\usepackage{balance} 
\usepackage{multirow}
\usepackage{subcaption}
\usepackage{xcolor}
\usepackage{amsmath}

\usepackage{enumitem}
\usepackage{color}
\usepackage{amsfonts}

\usepackage{bm}
\newtheorem{assumption}{Assumption}

\usepackage{booktabs} 
\usepackage{multirow} 
\usepackage{makecell} 
\usepackage{colortbl} 

\usepackage[noend,ruled,linesnumbered]{algorithm2e}  

\usepackage{caption} 

\usepackage[T1]{fontenc} 

\setcopyright{ifaamas}
\acmConference[AAMAS '22]{Proc.\@ of the 21st International Conference
on Autonomous Agents and Multiagent Systems (AAMAS 2022)}{May 9--13, 2022}
{Online}{P.~Faliszewski, V.~Mascardi, C.~Pelachaud,
M.E.~Taylor (eds.)}
\copyrightyear{2022}
\acmYear{2022}
\acmDOI{}
\acmPrice{}
\acmISBN{}

\acmSubmissionID{171}

\title{Mis-spoke or mis-lead: Achieving Robustness in Multi-Agent Communicative Reinforcement Learning}

\author{Wanqi Xue}
\affiliation{
  \institution{Nanyang Technological University}
  \country{Singapore}}
\email{wanqi001@e.ntu.edu.sg}

\author{Wei Qiu$^{*}$}\thanks{$^{*}$ Corresponding author}
\affiliation{
  \institution{Nanyang Technological University}
  \country{Singapore}}
\email{qiuw0008@e.ntu.edu.sg}

\author{Bo An}
\affiliation{
  \institution{Nanyang Technological University}
  \country{Singapore}}
\email{boan@ntu.edu.sg}

\author{Zinovi Rabinovich}
\affiliation{
  \institution{Nanyang Technological University}
  \country{Singapore}}
\email{zinovi@ntu.edu.sg}

\author{Svetlana Obraztsova}
\affiliation{
  \institution{Nanyang Technological University}
  \country{Singapore}}
\email{lana@ntu.edu.sg}

\author{Chai Kiat Yeo}
\affiliation{
  \institution{Nanyang Technological University}
  \country{Singapore}}
\email{asckyeo@ntu.edu.sg}

\begin{abstract}
{\color{black}Recent studies in multi-agent communicative reinforcement learning (MACRL) have demonstrated that multi-agent coordination can be greatly improved by allowing communication between agents. Meanwhile, adversarial machine learning (ML) has shown that ML models are vulnerable to attacks. Despite the increasing concern about the robustness of ML algorithms, how to achieve robust communication in multi-agent reinforcement learning has been largely neglected. In this paper, we systematically explore the problem of adversarial communication in MACRL. Our main contributions are threefold. First, we propose an effective method to perform attacks in MACRL, by learning a model to generate optimal malicious messages. Second, we develop a defence method based on message reconstruction, to maintain multi-agent coordination under message attacks. Third, we formulate the adversarial communication problem as a two-player zero-sum game and propose a game-theoretical method $\mathfrak{R}$-MACRL to improve the worst-case defending performance.
Empirical results demonstrate that many state-of-the-art MACRL methods are vulnerable to message attacks, and our method can significantly improve their robustness.}
\end{abstract}


\newcommand{\BibTeX}{\rm B\kern-.05em{\sc i\kern-.025em b}\kern-.08em\TeX}

\begin{document}


\pagestyle{fancy}
\fancyhead{}


\maketitle 

{\color{black}
\section{Introduction}\label{sec:intro}
Cooperative multi-agent reinforcement learning (MARL) has achieved remarkable success in a variety of challenging problems, such as urban systems~\cite{singh2020hierarchical}, coordination of robot swarms~\cite{huttenrauch2017guided} and real-time strategy video games~\cite{vinyals2019grandmaster}.} 
{\color{black} To tackle the problems of scalability and non-stationarity in MARL, the framework of centralized training with decentralized execution (CTDE) is proposed~\cite{oliehoek2008optimal,kraemer2016multi}, where decentralized policies are learned in a centralized manner so that they can share experiences, parameters, etc., during training. Despite the advantages, CTDE-based methods still perform unsatisfactorily in scenarios where multi-agent coordination is necessary, mainly due to partial observability in decentralized execution. To mitigate partial observability, many multi-agent communicative reinforcement learning (MACRL) methods have been proposed, which allow agents to exchange information such as private observations, intentions during the execution phase.
MACRL greatly improves multi-agent coordination in a wide range of tasks~\cite{foerster2016learning,sukhbaatar2016learning,jiang2018learning,das2019tarmac,ndq2020iclr,wang2020imac,kimcommunication}.

Meanwhile, adversarial machine learning has received extensive attention~\cite{huang2011adversarial,barreno2010security}. Adversarial machine learning demonstrates that machine learning (ML) models are vulnerable to manipulation~\cite{goodfellow2017attacking,Gleave2020Adversarial}. As a result, ML models often suffer from performance degradation when under attack. For the same reason, many practical applications of ML models are at high risk. For instance, researchers have shown that, by placing a few small stickers on the ground at an intersection, self-driving cars can be tricked into making abnormal judgements and driving into the opposite lane~\cite{tesla}. Maliciously designed attacks on ML models can have serious consequences.

Unfortunately, despite great importance, adversarial problems, especially adversarial inter-agent communication problems, remain largely uninvestigated in MACRL. Blumenkamp et al. show that, by introducing random noise in communication, agents are able to deceive their opponents in competitive games~\cite{blumenkamp2020emergence}. However, the attacks are not artificially designed and therefore inefficient. Besides, cooperative cases, where communication is more crucial, are neglected. They also fail to propose an effective defence, but merely retrain the models to adapt to the attacks. Mitchell et al. propose
to generate weights of Attention models~\cite{vaswani2017attention} through Gaussian process for defending against random attacks in attention-based MACRL~\cite{mitchell2020gaussian}. However, the applicability of this approach is unsatisfactory, being limited to attention-based MACRL, and its performance on maliciously designed attacks is unclear.

In this paper, we systematically explore the problem of adversarial communication in MACRL, where there are malicious agents that attempt to disrupt multi-agent cooperation by manipulating messages.
Our contributions are in three aspects. First, we propose an effective learning approach to model the optimal attacking scheme in MACRL. 
Second, to defend against the adversary, we propose a two-stage message filter which works by first detecting the malicious message and then recovering the message. The defending scheme can greatly maintain the coordination of agents under message attacks. Third, to address the problem that the message filter can be exploited by learnable attackers, we formulate the attack-defense problem as a two-player zero-sum game and propose $\mathfrak{R}$-MACRL, based on a game-theoretical framework Policy-Space Response Oracle (PSRO)~\cite{lanctot2017unified,muller2019generalized}, to approximate a Nash equilibrium policy in the adversarial communication game. $\mathfrak{R}$-MACRL improves the defensive performance under the worst case and thus improves the robustness.
Empirical experiments demonstrate that many state-of-the-art MACRL algorithms are vulnerable to attacks and our method can significantly improve their robustness.}

\section{Preliminaries and Related Work}\label{sec:relatedworks}

\textbf{Multi-Agent Communicative Reinforcement Learning.}
There has been extensive research on encouraging communication between agents to improve performance on cooperative or competitive tasks. 
Among the recent advances, some design communication mechanisms to address the problem of when to communicate~\cite{kim2019learning,singh2018learning,jiang2018learning}; other lines of works, e.g., TarMac~\cite{das2019tarmac}, focus on who to communicate. These works determine the two fundamental elements in communication, i.e., the message sender and the receiver. Apart from the two elements, the message itself is another element which is crucial in communication, i.e., what to communicate: Jaques et al. propose to maximize the social influence of messages~\cite{jaques2019social}. Kim et al. encode messages such that they contain the intention of agents~\cite{kimcommunication}. Some other works learn to send succinct messages to meet the limitations of communication bandwidth~\cite{ndq2020iclr,zhang2020tmc,wang2020imac}. {\color{black}Despite significant progress in MACRL, if some agents are adversarial and send maliciously designed messages, multi-agent coordination will rapidly disintegrate as these messages propagate.

\noindent \textbf{Adversarial Training.} Adversarial training is a prevalent paradigm for training robust models to defend against potential attacks~\cite{goodfellow2017attacking,szegedy2014intriguing}.
Recent literature has considered two types of attacks~\cite{carlini2017towards,papernot2017practical,tramer2018ensemble}: black-box attack and white-box attack. In black-box attack, the attacker does not have access to information about the attacked deep neural network (DNN) model; whereas in white-box attack, the attacker has complete knowledge, e.g., the architecture, the parameters and potential defense mechanisms, about the DNN model. We consider the black-box attack in our problem formulation, because the setting of the white-box attack is too idealistic and may not be applicable to many realistic adversarial scenarios. 
In adversarial training, the attacker tries to attack a DNN by corrupting the input via $\ell_{p}$-norm ($p \in \{1, 2, \infty\}$) attack~\cite{goodfellow2017attacking}. The attacker carefully generates artificial perturbations to manipulate the input of the model. In doing so, the DNN will be fooled into making incorrect predictions or decisions. The attacker finds the optimal perturbation $\delta$ by optimizing:
\begin{equation}
\underset{\delta}\max\mathcal{L}_{\text{predict}}\left(f(\boldsymbol{x}), f(\boldsymbol{x}+\delta) \right) \quad \text { s.t. } \quad  \min\mathcal{L}_{\text{norm}}\left(\boldsymbol{x}, \boldsymbol{x}+\delta \right) \nonumber
\end{equation}
where $\boldsymbol{x}$ is the input, $f$ is the DNN model, $\mathcal{L}_{\text{predict}}$
is a metric to measure the distance between the outputs of the DNN model w/ and w/o being attacked, $\mathcal{L}_{\text{norm}}$ is used to measure that for the inputs.

\noindent \textbf{Adversarial Reinforcement Learning (RL).} Recent advances in adversarial machine learning motivate researchers to investigate the adversarial problem in RL~\cite{Gleave2020Adversarial,lin2017tactics,xu2021transferable}.
SA-MDP~\cite{zhang2020robust} characterizes the problem of decision making under adversarial attacks on state observations.
Lin et al. propose two tactics of attacks, i.e., the strategically-timed attack and the enchanting attack, which attack by injecting noise to states and luring the agent to a designated target~\cite{lin2017tactics}. Gleave et al. consider the problem of taking adversarial actions that change the environment and consequentially change the observation of agents~\cite{Gleave2020Adversarial}. ATLA~\cite{zhang2021robust} propose to train the optimal adversary to perturb state observations and improve the worst-case agent reward. The settings of these works are different from ours: we consider the multi-agent scenario and restrict the attacking approach to adversarial messages, which makes the detection of anomalies difficult. Tu et al. propose to attack on multi-agent communication~\cite{tu2021adversarial}. However, their focus is on the representation-level, whereas we focus on the policy-level. Recently, there are some works considering a similar setting as ours~\cite{blumenkamp2020emergence,mitchell2020gaussian}. However, they either focus on random attacks in specific competitive games or the defence 
of specific communication methods.} 

\begin{figure*}
    \centering
    \includegraphics[width=0.85\linewidth]{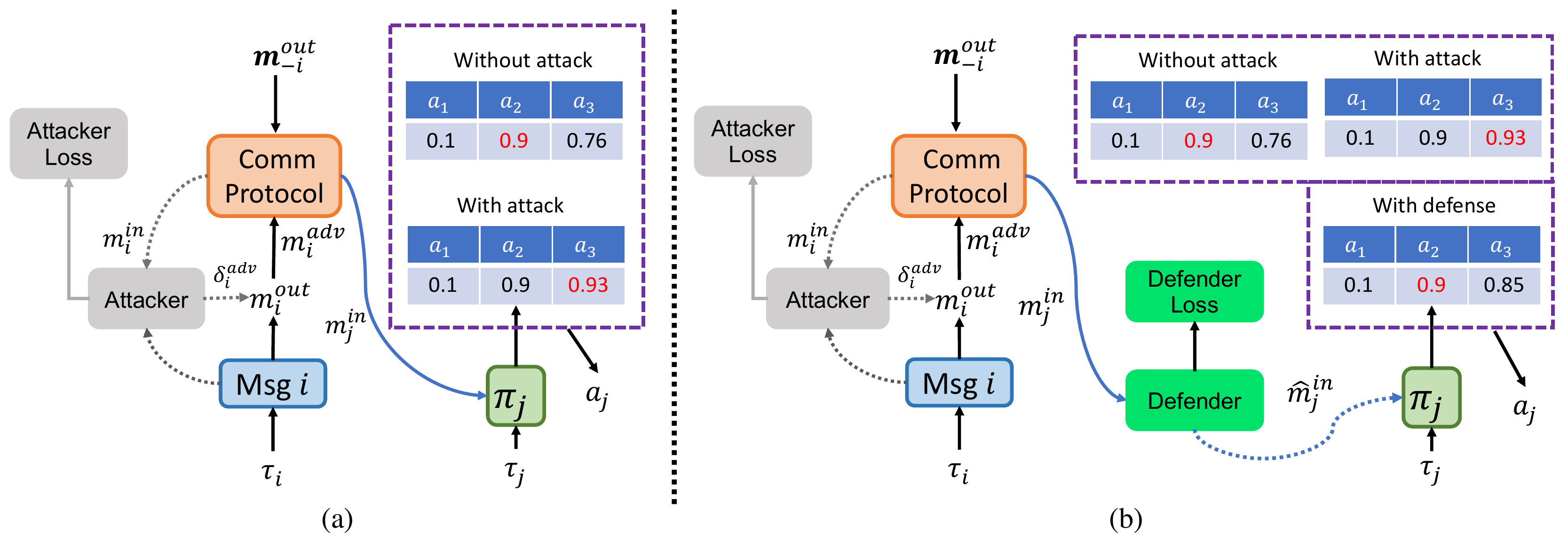}
    \caption{{\color{black}General framework of attack (left) and defence (right) in MACRL. \textit{Attack:} The agent $i$, taken over by the attacker, generates malicious message $m_i^{adv}$ to disrupt multi-agent cooperation. The incoming message $m_j^{in}$ and estimated Q-values of the benign agent $j$ will be affected due to $m_i^{adv}$, which may lead to incorrect decisions. \textit{Defence:} A learnable defender is cascaded to the communication protocol module, which is used to reconstruct the contaminated message ($m_j^{in}$ to $\hat{m}_j^{in}$). The benign agent $j$ estimates Q-values based on $\hat{m}_j^{in}$, which can reduce the probability of selecting sub-optimal actions.}}
    \label{fig:ndq_attack_n_defend}
\end{figure*}


\section{Achieving Robustness in MACRL}

In this section, we propose our method for achieving robustness in MACRL.
We begin by proposing the problem formulation for adversarial communication in MACRL. Then, we introduce the method to build the optimal attacking scheme in MACRL. Next, we propose a learnable two-stage message filter to defend against possible attacks. Finally, we propose to formulate the attack-defense problem as a two-player zero-sum game, {\color{black} and design a game-theoretic method $\mathfrak{R}$-MACRL} to approximate a Nash equilibrium policy for the defender.  {\color{black}In this way, we can improve the worst-case performance of the defender and thus enhance robustness.}

\subsection{Problem Formulation: Adversarial Communication in MACRL}\label{sec:formulation}
{\color{black}An MACRL problem can be modeled by \textit{Decentralised Partially Observable Markov Decision Process with Communication} (\textit{Dec-POMDP-Com})~\cite{oliehoek2016concise}, which is defined by a tuple $\langle \mathcal{S}, \mathcal{M}, \mathcal{A},\mathcal{P},R, \Upsilon,O,C,\mathcal{N},\gamma\rangle$. $\mathcal{S}$ denotes the state of the environment and $\mathcal{M}$ is the message space. Each agent $i \in \mathcal{N} := \{1,...,N\} $ chooses an action $a_i \in \mathcal{A}$ at a state $\bm{s} \in \mathcal{S}$, giving rise to a joint action vector, $\bm{a} := [a_i]_{i=1}^N \in \mathcal{A}^N$. $\mathcal{P}(\bm{s}'|\bm{s},\bm{a}):\mathcal{S} \times \mathcal{A}^N \times \mathcal{S} \mapsto \mathcal{P}(\mathcal{S})$ is a Markovian transition function. Every agent shares the same joint reward function $R(\bm{s},\boldsymbol{a}): \mathcal{S} \times \mathcal{A}^N \mapsto \mathbb{R} $, and $\gamma \in [0,1)$ is the discount factor. Due to partial observability, each agent has individual partial observation $\upsilon \in \Upsilon$, according to the observation function $O(\bm{s},i): \mathcal{S} \times \mathcal{N} \mapsto \Upsilon.$ Each agent holds two policies, i.e., action policy $p_i(a_i|\tau_i, m_i^{in}) : \mathcal{T} \times \mathcal{M} \times \mathcal{A} \mapsto [0,1]$ and message policy  $v_i(m_i^{out}|\tau_i, m_i^{in}) : \mathcal{T} \times \mathcal{M} \times \mathcal{M} \mapsto [0,1]$,
both of which are conditioned on the action-observation history $\tau_i \in \mathcal{T} := (\Upsilon \times \mathcal{A})$ and incoming messages $m_i^{in}$ aggregated by the communication protocol $C(m_i^{in}|\bm{m}^{out}, i): \mathcal{M}^{|\mathcal{N}|} \times \mathcal{N} \times \mathcal{M} \mapsto [0,1]$.
}

In adversarial communication where there are $N_{adv}$ malicious agents, we assume that each malicious agent holds the third private adversarial policy, $\xi(\delta_i^{adv}|\tau_{i}, m_i^{in}, m_i^{out}) : \mathcal{T} \times \mathcal{M} \times \mathcal{M} \times \mathcal{M} \mapsto [0,1]$, which generates adversarial message $\delta_i^{adv}$ based on its action-observation history $\tau_{i}$, received messages $m_i^{in}$ and the message $m_i^{out}$ intended to be sent. 
Malicious agents could send messages by convexly combining their original messages with adversarial messages, i.e., $m_i^{adv}=(1-\omega) \times m_i^{out} + \omega \times \delta_i^{adv}$, or  simply summing up the messages, {\color{black}i.e., $m_i^{adv}=m_i^{out}+\delta_i^{adv}$}. To reduce the likelihood of being detected, apart from the adversarial policy $\xi$, malicious agents strictly follow their former action policy and message policy, trying to behave like benign agents. 
Fig. \ref{fig:ndq_attack_n_defend}(a) presents the overall attacking procedure. The agent $i$ (attacker) is malicious and tries to generate adversarial message $m_i^{adv}$ to disrupt cooperation. The adversarial message sent by agent $i$ together with normal messages sent by other agents (denoted by $\bm{m}_{-i}^{out}$) are processed by the communication protocol (algorithm-related), generating a contaminated incoming message $m_j^{in}$ for a benign agent $j$. From agent $j$'s perspective, under such attack, the estimated Q-values will change. {\color{black} If the action with the highest Q-value shifts, agent $j$ will make incorrect decisions, leading to suboptimality.}
To perform an effective attack in MACRL, we propose to optimize the adversarial policy $\xi$ by minimizing the joint accumulated rewards, i.e., $\min_{\xi} \mathbb{E}\left[\sum_{t=0}^{\infty} \gamma^{t} r_{t} \right]$. We make two assumptions to make the adversarial communication problem both practical and tractable.

\begin{assumption}
(Byzantine Failure~\cite{bf}) {\color{black}Agents have imperfect information on who are malicious.} 
\end{assumption}

\begin{assumption}
(Concealment) {\color{black}Malicious agents do not communicate or cooperate with each other when performing attacks in order to be covert.}
\end{assumption}

To defend against the attacker, as in Fig. \ref{fig:ndq_attack_n_defend}(b), we propose to cascade a learnable defender to the communication protocol module. {\color{black}The defender performs a transformation from $m_j^{in}$ to $\hat{m}_j^{in}$ to reconstruct the contaminated messages, to avoid distributing the contaminated message $m_j^{in}$ directly to agent $j$. With such transformation, the benign agent $j$ can estimate the Q-value for each action more properly and reduce the probability of selecting sub-optimal actions.}

\subsection{Learning the Attacking Scheme}\label{sec:attacker}
{\color{black}
To model the attack scheme in adversarial communication, we use a deep neural network (DNN) $f_\mu$, parameterized by $\theta_\mu$, to generate adversarial messages for a malicious agent. The adversarial policy $\xi$ is a multivariate Gaussian distribution whose parameters are determined by the DNN $f_\mu$, i.e., $\xi=\mathcal{N}(f_\mu(\cdot|\tau, m^{in}, m^{out};\theta_{d}),\Lambda)$ where  $\Lambda$ is a fixed covariance matrix.
The reason of using Gaussian distribution as the prior is that it is the maximum entropy distribution under constraints of mean and variance. Each malicious agent generates adversarial messages by sampling from its adversarial policy. The optimization objective of the adversarial policy is to minimize the accumulated team rewards subject to a constraint on the distance between $m^{out}$ and $m^{adv}$. We utilize Proximal Policy Optimization (PPO)~\cite{schulman2017proximal} to optimize the adversarial policy by maximizing the following objective:
\begin{small}
\begin{multline}
     \mathcal{J}_\xi(\theta_{\mu})=
    \mathbb{E}_{(\tau,m^{in},m^{out},\delta^{adv})}[\min(
   \rho \cdot A^{\xi}, \operatorname{clip}\left(\rho, 1\pm\epsilon \right)\cdot A^{\xi}
    )] \\ -\alpha \cdot \mathbb{E}_{(m^{out},\delta^{adv})}[(m^{out}-m^{adv})^2] 
    \label{eq:learn_attacker}
\end{multline}
\end{small}where $\rho=\xi(\delta^{adv}|\tau, m^{in}, m^{out})/\xi^{old}(\delta^{adv}|\tau, m^{in}, m^{out})$,  $\xi^{old}$ is the policy in the previous learning step, $\epsilon$ is the clipping ratio, and $A^{\xi}(\delta^{adv},\tau, m^{in}, m^{out})$ is the advantage function. Let $e=\langle\tau, m^{in}, m^{out}\rangle$ denotes the input of the value function and we define the reward of the attacker to be negative team reward, then $A^\xi(\delta^{adv},e)=-r+V(e')-V(e)$ where $V$ is the value function, $e'$ denotes the next step state, and $r$ is the immediate reward. We learn the value function $V(e)$ by minimizing $\mathbb{E}[(V(e)+\sum_{t} \gamma^{t} r_{t})^2]$.
}

\subsection{Defending against Adversarial Communication}\label{sec:defence}
\begin{figure}
\centering
\includegraphics[width=0.65\linewidth]{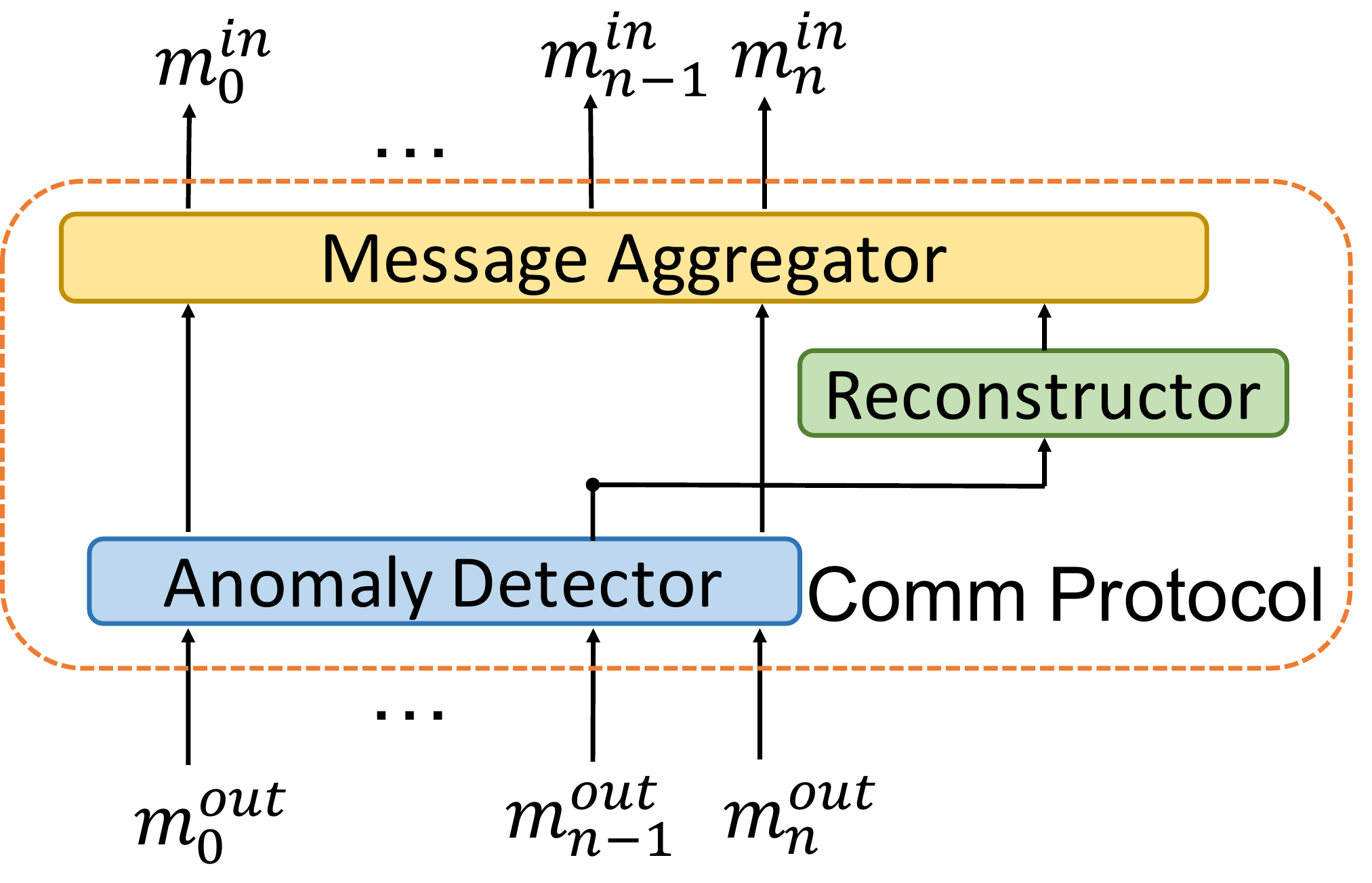}
\caption{{\color{black}Structure of the communication protocol with the message filter (defender). The anomaly detector and the reconstructor are designed to determine and recover the potential malicious messages, respectively.}}\label{fig:defender}
\vspace{-0.5cm}
\end{figure}

We can train a DNN to model the attack scheme in adversarial communication. However,
the defence of that is non-trivial because i) benign agents have no prior knowledge on which agents are malicious; ii) the malicious agents can inconsistently pretend to be cooperative or non-cooperative to avoid being detected; and iii) recovering useful information from the contaminated messages is difficult. To address these challenges, we design a two-stage message filter for the communication protocol to perform defence. {\color{black}The message filter works by first determining the messages that are likely to be malicious and then recovering the potential malicious messages before distributing them to the corresponding agents.} 
As in Fig. \ref{fig:defender}, the message filter $\zeta(h_d,g_r)$ consists of an anomaly detector $h_d$ and a message reconstructor $g_r$. The anomaly detector, parameterized by $\theta_{d}$, outputs the probability that each incoming message needs to be recovered, {\color{black}i.e., $h_d(\cdot|m_i^{out};\theta_{d}):\mathcal{M} \mapsto \Psi$, where $\Psi$ denotes a binomial distribution, $m_i^{out}$ denotes the message sent by agent $i$. We perform sampling from the generated distributions to determine the anomaly messages $\bm{x} \sim  h_d(\cdot|\bm{m}^{out};\theta_{d})$ \footnote{{\color{black}We abuse $h_d(\cdot|\bm{m}^{out};\theta_{d})$ to represent that messages $m^{out}_i\in \bm{m}^{out}$ are fed into $h_d$ in batch.}}. Here, $\bm{x}$ is an indicator that records whether a message is predicted to be malicious or not. The predicted malicious messages $\bm{\dot{m}}$ are recovered by the reconstructor $g_r(\cdot|\dot{m}_i;\theta_r) : \mathcal{M}\mapsto\mathcal{M}$ separately.} The recovered message and other normal messages are aggregated and determine the messages that each agent will receive. 

The optimization objective of the message filter is to maximize the joint accumulated rewards under attack, i.e., $\max_{\zeta} \mathbb{E}\left[\sum_{t=0}^{\infty} \gamma^{t} \tilde{r_{t}} |\xi \right]$, where $\tilde{r_{t}}$ is the team reward after performing the defending strategy. 
We utilize PPO to optimize the strategy. To mitigate the local optima induced by unguided exploration in large message space, {\color{black}we introduce a regularizer which is optimized under the guidance of the ground-truth labels of messages (whether they are malicious). For the reconstructor,} we train it by minimizing the distance between the messages before and after being attacked. To improve the tolerance of the reconstructor to errors made by the detector, apart from malicious messages, we also use benign messages as training data, in which case the reconstructor is an identical mapping. Formally, the two-stage message filter is optimized by maximizing the following function:{\color{black}
\begin{small}
\begin{multline}
\mathcal{J}_\zeta \left(\theta_d,\theta_r\right)=
\mathbb{E}_{(\bm{m}, \bm{x})} \left[\sum_i^{|\bm{m}|} \min \left(\rho_i \cdot A^{h_d}_i, \operatorname{clip}\left(\rho_i, 1\pm\epsilon\right)\cdot A^{h_d}_i\right)\right] \\
+\beta_1 \cdot \mathbb{E}_{\bm{m}}\left[\bm{\hat{y}}\cdot \log( h_d(\cdot|\bm{m};\theta_{d})) \right] \\
-\beta_2 \cdot \mathbb{E}_{\bm{m}}\left[ (\bm{\hat{m}}-g_r(\cdot|\bm{m};\theta_r))^2\right]
\label{eq:learn_defender}
\end{multline}
\end{small}where $\rho_i = h_d(x_i|m_i;\theta_d)/h_d^{old}(x_i|m_i;\theta_d)$, $A^{h_d}_i(x_i,m_i)$ is the advantages which are estimated similarly as in the attack, 
$\bm{\hat{y}}$ is the one-hot ground-truth labels of messages, $\bm{\hat{m}}$ is the messages that have not been attacked.} $\beta_1$, $\beta_2$ and $\epsilon$ are hyperparameters.

\subsection{\textcolor{black}{Achieving Robust MACRL}}\label{sec:r_macrl}
{\color{black}
Despite the defensive message filter, the effectiveness of the defence system can rapidly decrease if malicious agents are aware of the defender and adjust their attack schemes. 
To mitigate this, we formulate the attack-defence problem as a two-player zero-sum game (malicious agents are controlled by the attacker) and improve the performance of the defender in the worst case. We define the adversarial communication game by a tuple $\langle\Pi$, $U\rangle$, where $\Pi=\langle\Pi_\xi, \Pi_\zeta\rangle$ is the joint policy space of the players ($\Pi_\xi$ and $\Pi_\zeta$ denote the policy space of the defender and the attacker respectively), $U : \Pi \mapsto \mathbb{R}^2$ is the utility function which is used to calculate utilities for the attacker and the defender given their joint policy $\pi=\langle\pi_\xi,\pi_\zeta \rangle \in\Pi$. The utility of the defender is defined as the expected team return. The adversarial communication game is zero-sum, i.e., the utility of the defender $U_{\zeta}(\pi)$ must be the negative of the utility of the attacker $U_{\xi}(\pi)$ for $\forall \pi \in \Pi$. The solution concept of the adversarial communication game is Nash equilibrium (NE) and we can approach an NE by optimizing the following MaxMin objective:
\begin{small}
\begin{equation}
    \mathcal{J}_{\xi,\zeta} =\max_{\pi_{\zeta}\in\Pi_{\zeta}}U_{\zeta}(\operatorname{Br}(\pi_\zeta) ,\pi_{\zeta})
    =\max_{\pi_{\zeta}\in\Pi_{\zeta}} \min_{\pi_{\xi}\in\Pi_{\xi}} \mathbb{E}\left[\sum\nolimits_{t=0}^{\infty} \gamma^{t} \tilde{r_{t}} |\pi_{\xi},\pi_{\zeta}\right]
\end{equation}
\end{small}where $\operatorname{Br}(\pi_\zeta)$  is the best response policy of the defender, i.e., $\operatorname{Br}(\pi_\zeta)=\arg\min_{\pi_{\xi}}U_\zeta(\pi_{\xi},\pi_\zeta)$.}

\begin{figure}
\centering
\includegraphics[width=0.5\linewidth]{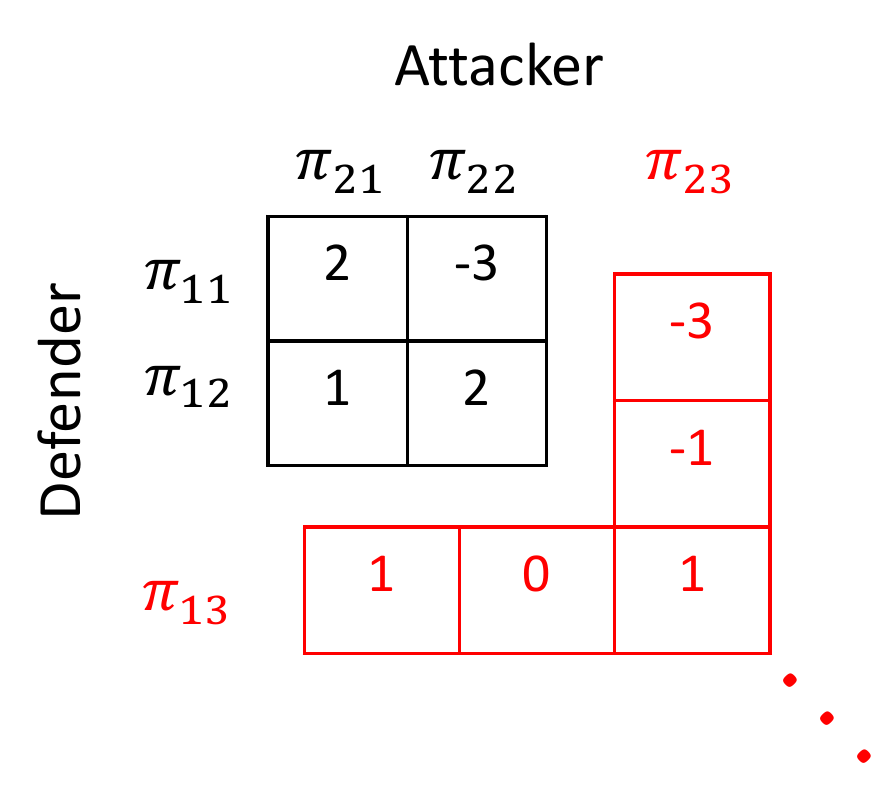}
\caption{Workflow of $\mathfrak{R}$-MACRL.}\label{fig:missing_entries}
\end{figure}

We propose $\mathfrak{R}$-MACRL to optimize the objective in the adversarial communication game. $\mathfrak{R}$-MACRL is developed based on Policy-Space Response Oracle (PSRO)~\cite{lanctot2017unified}, a deep learning-based extension of the classic Double Oracle algorithm~\cite{mcmahan2003planning}. 
The workflow of $\mathfrak{R}$-MACRL is depicted in Fig.~\ref{fig:missing_entries}. 
{\color{black}At each iteration, $\mathfrak{R}$-MACRL keeps a population (set) of policies $\Pi^t\subset\Pi$, e.g., $\Pi^t_{\zeta}=(\pi_{11},\pi_{12})$ and $\Pi^t_{\xi}=(\pi_{21},\pi_{22})$. We can evaluate the utility table $U(\Pi^{t})$ for the current iteration and calculate a Nash equilibrium $\pi^t_*=\langle\Delta(\Pi^t_{\xi}), \Delta(\Pi^t_{\zeta})\rangle$ for the game restricted to policies in $\Pi^t$ by, for example, linear programming, replicator dynamics, etc. $\Delta$ denotes the meta-strategy which is an arbitrary categorical distribution. Next, we calculate the best response policy  $\operatorname{Br}(\pi^t_{*,-i})$ to the NE in this restricted game for each player $i$ (player $-i$ denotes the opponent of player $i$, $i\in\{\xi,\zeta\}$), e.g., $\pi_{13}=\operatorname{Br}(\pi^t_{*,\xi})$ and $\pi_{23}=\operatorname{Br}(\pi^t_{*,\zeta})$. We extend the population by adding the best response policies to the policy set: $\Pi^{t+1}_i=\Pi^{t}_i\cup\operatorname{Br}(\pi^t_{*,-i})$ for $i\in\{\xi,\zeta\}$. After extending the population, we complete the utility table $U(\Pi^{t+1})$ and perform the next iteration.} The algorithm converges if the best response policies are already in the population.
Practically, $\mathfrak{R}$-MACRL approximates the utility function $U(\cdot)$ by having each policy in the population $\Pi^t_i$ playing with each other policy in $\Pi^t_{-i}$ and tracking the average utilities. The approximation of the best response policies is performed by 
maximizing $\mathcal{J}_{\xi}(\theta_\mu|\pi^t_{*,\zeta})$ and $\mathcal{J}_{\zeta}(\theta_d,\theta_r|\pi^t_{*,\xi})$ 
for the attacker and the defender, where the full expressions of the two objectives are described in Eq.~\ref{eq:learn_attacker} and Eq.~\ref{eq:learn_defender}, respectively.

\begin{algorithm}[t]
\textbf{Input:} initial policy sets for both players $\Pi^0$, maximum number of iterations $T$;\\
Empirically estimate utilities $U(\Pi^0)$ for each joint policy $\pi \in \Pi^0$\;\label{algo:meta-solver}
Initialize mixed strategies for both players $\pi^0_{*,i} = \textsc{Uniform}(\Pi^0_i)$\ for $i\in\{{\xi,\zeta}\}$\;\label{algo:uniform}
Initialize number of iterations $t=0$;\\
\While{not converge and $t\leq T$}{
  \For{player $i \in \{\xi, \zeta\}$}{
    Train $\pi^{t+1}_i$ by letting it exploit $\pi^t_{*,-i}$;\label{algo:algo1_exploit}\\
  Extend policy set $\Pi^{t+1}_i = \Pi_i^t \cup \{ \pi_i^{t+1} \}$\label{algo:algo1_extend};
  }
  Check convergence and $t=t+1$;\\\label{algo:convergence}
  Estimate missing entries in $U(\Pi^{t+1})$\label{algo:algo1_compute} \;
  Compute mixed strategies $\pi_*^{t+1}$ by solving the matrix game defined by $U(\Pi^{t+1})$\label{algo:algo1_nash} ;
}
\textbf{Output:} mixed strategies $\pi_{*}$ for both players \;
\caption{$\mathfrak{R}$-MACRL}\label{alg:psro}
\end{algorithm}
{\color{black}The overall algorithm of $\mathfrak{R}$-MACRL is presented in Algorithm \ref{alg:psro}. We first initialize the meta-game and the mixed strategies (line \ref{algo:meta-solver} and line \ref{algo:uniform}). Then, at each step, we train  the attacker and the defender alternately by letting them best respond to their corresponding opponent (line \ref{algo:algo1_exploit}). Next, the learned new policy $\pi^{t+1}_{i}$ is added to the policy set $\Pi^{t+1}_{i}$ for the two players respectively (line \ref{algo:algo1_extend}). After extending the policy set,  we can check the convergence (line \ref{algo:convergence}) and calculate the missing entries in $U(\Pi^{t+1})$ (line \ref{algo:algo1_compute}). Finally, we solve the new meta-game (line \ref{algo:algo1_nash}) and repeat the iteration.}

\section{Experiments}\label{sec:experiments}
We conduct extensive experiments on a variety of state-of-the-art MACRL algorithms to answer:
\textbf{Q1:} \textit{Are MACRL methods vulnerable to message attacks and whether the two-stage message filter is able to consistently recover multi-agent coordination?} \textbf{Q2:} \textit{Whether $\mathfrak{R}$-MACRL is able to improve the robustness of MACRL algorithms under message attacks?} \textbf{Q3:} \textit{Whether our method is able to scale to scenarios where there are multiple attackers?} \textbf{Q4:} \textit{Which components contribute to the performance of the method and how does the proposed method work?} 
We first categorize existing MACRL algorithms and then select representative algorithms to perform the evaluation. All experiments are conducted on a server with 8 NVIDIA Tesla V100 GPUs and a 44-core 2.20GHz Intel(R) Xeon(R) E5-2699 v4 CPU.

\subsection{Experimental Setting}
\label{sec:taxonomy}

MACRL methods are commonly categorized by whether they are implicit or explicit~\cite{oliehoek2016concise,ahilan2020correcting}. In implicit communication, the actions of agents influence the observations of the other agents.
Whereas in explicit communication, agents have a separate set of communication actions and exchange information via communication actions. In this paper, we focus on explicit communication because in implicit communication, to carry out an attack, the attacker's behaviour is often bizarre, making the attack trivial and easily detectable. We consider the following three realistic types of explicit communication:
\begin{itemize}
    \item \textbf{Communication with delay (CD):} 
    Communication in real world is usually not real-time. We can model this by assuming that it takes one or a few time steps for the messages being received by the targeted agents~\cite{sukhbaatar2016learning}.
    \item \textbf{Local communication (LC):} Messages sometimes cannot be broadcast to all agents due to communication distance limitations or privacy concerns. Therefore, agents need to learn to communicate locally, affecting only those agents within the same communication group~\cite{das2019tarmac,bohmer2020dcg}.
    \item \textbf{Communication with cost or limited bandwidth (CC):} Agents should avoid sending redundant messages because communication in real world is costly and communication channels often have a limited bandwidth~\cite{ndq2020iclr,zhang2020tmc}.
\end{itemize}
Following the above taxonomy, we select some representative state-of-the-art algorithms to perform evaluation\footnote{Sweeping the whole list of algorithms for each category is impossible and unnecessary since there have been a lot of algorithms proposed and algorithms in each category usually share common characteristics.}. As in Table \ref{table:baselines}, we select CommNet~\cite{sukhbaatar2016learning}, TarMAC~\cite{das2019tarmac} and NDQ~\cite{ndq2020iclr} for CD, LC and CC respectively. A brief introduction to each of the chosen algorithms is provided in Appendix~\ref{app:intro_to_algo}. We evaluate in the following environments, which are similar to those in the paper that first introduced the algorithms:

\begin{table}
\centering 
    \begin{tabular}{{l}{l}{l}}
     \toprule
     Categories  & Methods & Environments\\
     \hline
        \textbf{CD} & CommNet~\cite{sukhbaatar2016learning} & Predator Prey (PP)~\cite{bohmer2020dcg} \\
        \textbf{LC} & TarMAC~\cite{das2019tarmac}  & Traffic Junction (TJ)~\cite{sukhbaatar2016learning}\\
        \textbf{CC} & NDQ~\cite{ndq2020iclr} & StarCraft II (SCII)~\cite{samvelyan19smac} \\
     \bottomrule
    \end{tabular}
  \captionof{table}{The chosen algorithms and environments.}\label{table:baselines}
  \vspace{-0.7cm}
\end{table}
\noindent\textbf{Predator Prey (PP).} There are 3 predators, trying to catch 6 prey in a $7 \times7$ grid. Each predator has local observation of a $3\times3$ sub-grid centered around it and can move in one of the 4 compass directions, remain still, or perform catch action. The prey moves randomly and is caught if at least two nearby predators try to catch it simultaneously. The predator will obtain a team reward at 10 for each successful catch and will be punished $p$ for each losing catch action. There are two tasks with $p=0$ and $p=-0.5$, respectively. A prey will be removed from the grid after it being caught. An episode ends after 60 steps or all preys have been caught.

\noindent\textbf{Traffic Junction (TJ).} In TJ, there are $N_{max}$ cars and the aim of each car is to complete its pre-assigned route. Each car can observe of a $3 \times 3$ region around it, but is free to communicate with other cars. The action space for each car at every time step is \{\texttt{gas}, \texttt{brake}\}. The reward function is $-0.01\tau+r_\textnormal{collision}$, where $\tau$ is the number of time steps since the activation of the car, and $r_\textnormal{collision} = -10$ is a collision penalty. Cars become available to be sampled and put back to the environment with new assigned routes once they complete their routes.

\noindent\textbf{StarCraft II (SCII).} We consider SMAC~\cite{samvelyan19smac} combat scenarios where the enemy units are controlled by StarCraft II built-in AI (difficulty level is set to hard), and each of the ally units is controlled by a learning agent. The units of the two groups can be asymmetric. The action space contains \texttt{no-op}, \texttt{move[direction]}, \texttt{attack[enemy id]}, and \texttt{stop}. Agents receive a globally shared reward which calculates the total damage made to the enemy units at each time step. We conduct experiments on four SMAC tasks: 3bane\_vs\_hM, 4bane\_vs\_hM, 1o\_2r\_vs\_4r and 1o\_3r\_vs\_4r. More details about the tasks are in Appendix~\ref{app:scii_map}.

For each of the tasks, we first train the chosen MACRL methods to obtain the action policy and the message policy for each agent. After that, we randomly select an agent to be malicious and train its adversarial policy to examine whether MACRL methods  are vulnerable to message attacks. Then we perform defence by training the message filter, during which the adversarial policy of the malicious agent is fixed but keeps working. To show that a single message filter is brittle and can be easily exploited if the attacker adapts to it, we freeze the learned message filter and retrain the adversarial policy. Finally, we integrate the message filter into the framework of $\mathfrak{R}$-MACRL and justify if $\mathfrak{R}$-MACRL is helpful to improve the robustness.
All experiments are carried out with five random seeds and results are shown with a 95\% confidence interval.

\subsection{Recovering Multi-Agent Coordination}\label{sec:attack_results}
We first evaluate the performance of our attacking method on the three selected MACRL algorithms. Then we try to recover multi-agent coordination for the attacked algorithms by applying the message filter.


\begin{figure}
    \centering
    \includegraphics[width=0.95\linewidth]{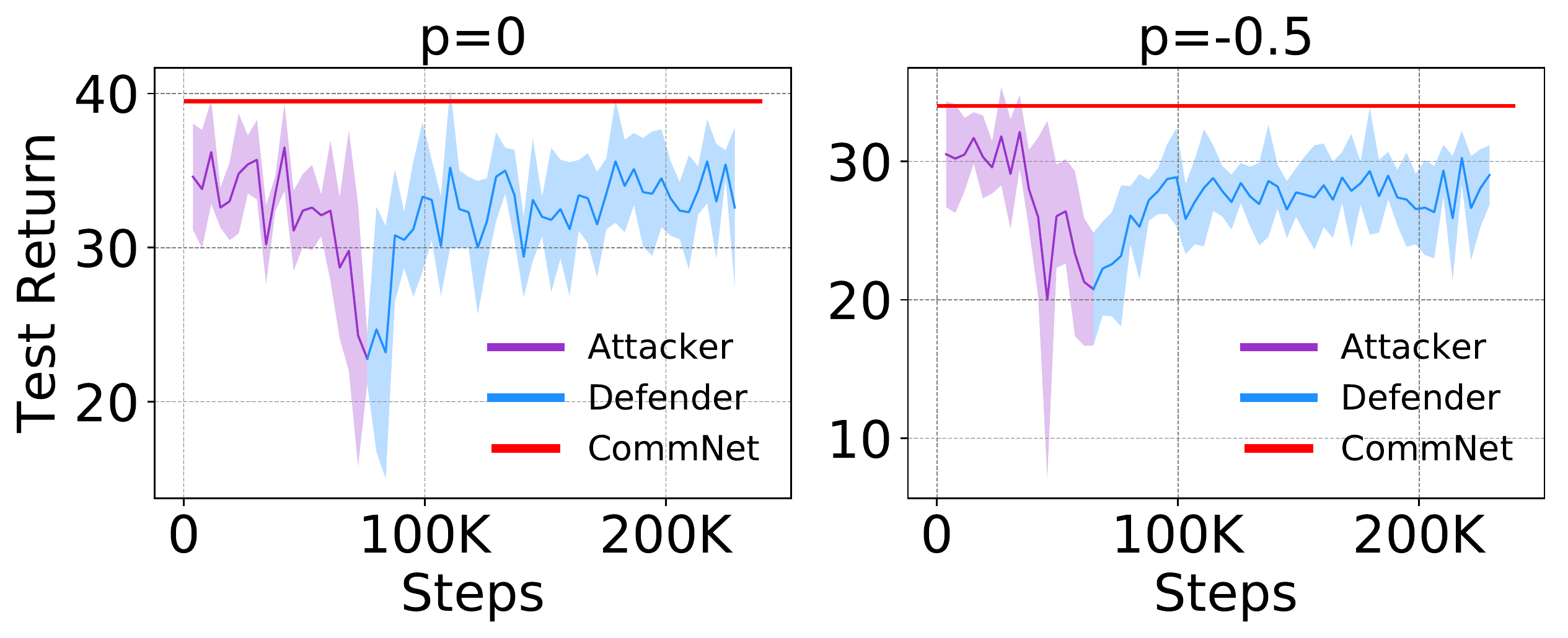}
    \vspace{-0.3cm}
    \captionof{figure}{Attack and defence on CommNet.}\label{fig:commnet_defense}
\end{figure}

\noindent\textbf{CommNet.} The experiments for CommNet are conducted on predator prey (PP)~\cite{bohmer2020dcg}. We set the punishment as $p=0$ and $p=-0.5$ to create two PP tasks with different difficulties.
As in Fig. \ref{fig:commnet_defense}, at the beginning of the attack, the performance of CommNet does not have obvious decrease, indicating that injecting random noise into the message is hard to disrupt agent coordination. As we gradually train the adversarial policy, there is a significant drop in the test return, with 40\% and 33\% decreases in the task of $p=0$ and $p=-0.5$, respectively. Multi-agent cooperation has been severely affected due to the malicious messages.
When the test return decreases to preset thresholds, i.e., 23 for $p=0$ and 20 for $p=-0.5$, we freeze the adversarial policy network and start to train the message filter. As shown in the blue curves in Fig \ref{fig:commnet_defense}, with the message filter, test return steadily approaches the converged value of CommNet (red line), indicating that the message filter can effectively recover multi-agent coordination under attack.




\begin{table}
\centering
\begin{tabular}{{l}{l}{l}}
\toprule
& Easy & Hard\\
\hline\hline
TarMAC & $99.9\pm0.1\%$ & $94.9\pm0.2\%$ \\
\hline
TarMAC w/ $\pi_{\xi}$ & $\boldsymbol{87.2\pm 4.68}\%$ & $\boldsymbol{88.75\pm 7.29}\%$ \\
TarMAC w/ $\pi_{\zeta}$ & $\boldsymbol{96.41\pm 1.38}\%$ & $\boldsymbol{93.23\pm 8.11\%}$ \\
\bottomrule
\end{tabular}
\caption{Attack and defence on TarMAC.}\label{table:results_tarmac_1}
\vspace{-0.7cm}
\end{table}

\noindent\textbf{TarMAC.} We conduct message attack and defence on the Traffic Junction (TJ) environment~\cite{sukhbaatar2016learning}. 
There are two modes in TJ, i.e., \texttt{easy} and \texttt{hard}.  The \texttt{easy} task has one junction of two one-way roads on a $7\times7$ grid with $N_{max}=5$. In the \texttt{hard} task, its map has four junctions of two-way roads on a $18\times18$ grid and $N_{max}=20$.
As shown in Table \ref{table:results_tarmac_1}, under attack, the success rate of TarMAC decreases in both the \texttt{easy} and the \texttt{hard} scenarios, demonstrating the vulnerability of TarMAC under malicious messages. After that, we equip TarMAC with the defender, then its performance improves considerably, demonstrating the merit of the message filter.

\begin{figure}
\centering
    \vspace{0.2cm}
  \includegraphics[width=0.95\linewidth]{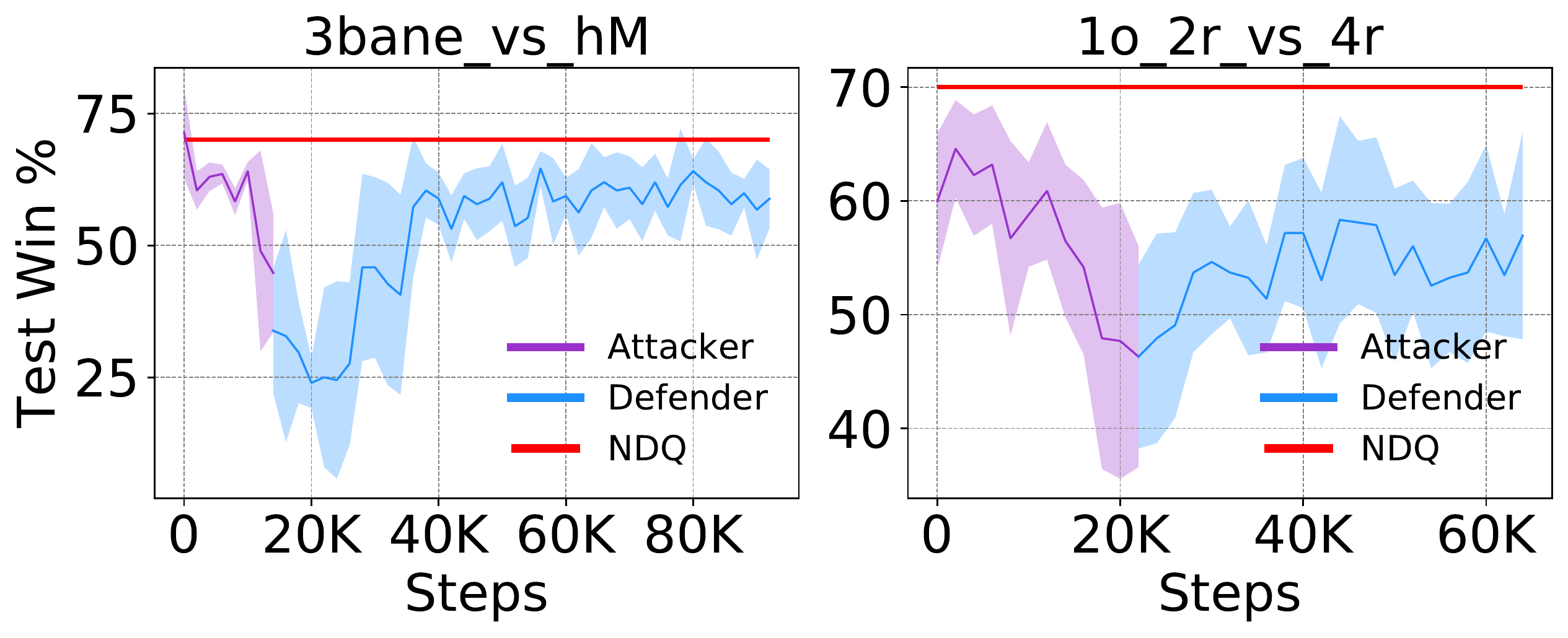}
  \vspace{-0.3cm}
  \caption{Attack and defence on NDQ.}
  \label{fig:ndq_results_1}
\end{figure}

\noindent\textbf{NDQ.} We further examine the adversarial communication problem in NQD. We perform evaluation on two StarCraft II Multi-Agent Challenge (SMAC)~\cite{samvelyan19smac} scenarios, i.e., $\operatorname{3bane\_vs\_hM}$ and $\operatorname{1o\_2r\_vs\_4r}$, in which the communication between cooperative agents is necessary~\cite{ndq2020iclr}. As presented in Fig. \ref{fig:ndq_results_1}, under attack, the test win rate of NDQ decreases dramatically, demonstrating that NDQ is also vulnerable to message attacks. After applying the message filter as the defender, we can find the test win rate quickly reaches to around $60\%$ in $\operatorname{3bane\_vs\_hM}$ and $55\%$ in $\operatorname{1o\_2r\_vs\_4r}$, demonstrating that, once again, our defence methods succeed in restoring multi-agent coordination under message attacks.


\begin{figure}
  \centering
  \vspace{-0.2cm}
  \includegraphics[width=0.95\linewidth]{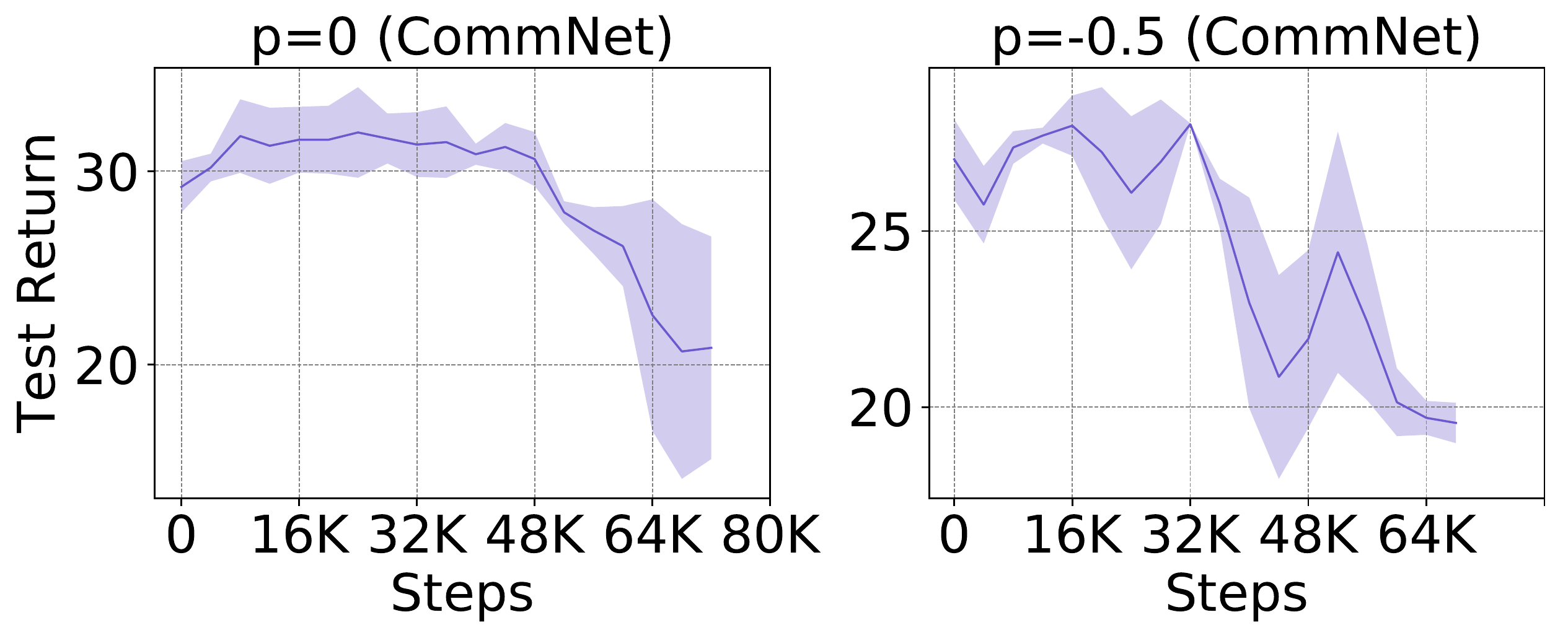}
  \vspace{-0.3cm}
  \caption{Exploiting the message filter in CommNet.}
  \label{fig:psro_adaptative_2}
\end{figure}
\subsection{Improving Robustness with $\mathfrak{R}$-MACRL}\label{sec:experiments_robustness_r_macrl}
We have demonstrated that many state-of-the-art MACRL algorithms are vulnerable to message attacks, and after applying the message filter, multi-agent coordination can be recovered. In this part, we integrate the message filter into the framework of $\mathfrak{R}$-MACRL and show that the robustness of MACRL algorithms can be significantly improved with $\mathfrak{R}$-MACRL.

\noindent\textbf{Exploiting the message filter.} We first show that a single message filter is brittle and can be easily exploited if the attacker adapts to it. We perform experiments on CommNet by freezing the message filter and retraining the adversarial policy. As depicted in Fig. \ref{fig:psro_adaptative_2}, the test return of the team gradually decreases as the training proceeds, with around 30\%  and 20\% decreases in the task of $p=0$ and $p=-0.5$ respectively. We also conduct experiments on NDQ and similar phenomenon is observed (see Appendix~\ref{appendix:more_results}). We conclude that even though the designed message filter is able to recover multi-agent cooperation under message attacks, its performance can degrade if faced with an adaptive attacker.

\begin{table}
    \centering
    \begin{tabular}{crcrcrcr}
        \toprule
        \multicolumn{2}{c}{Methods} &
        \multicolumn{2}{l}{Scenarios} &
        \multicolumn{2}{l}{$u^{\operatorname{\mathfrak{R}}}_{\zeta}$} &
        \multicolumn{2}{l}{$u^{vn}_{\zeta}$}\\
        \cmidrule(lr){1-2}
        \cmidrule(lr){3-4}
        \cmidrule(lr){5-8}
        
        \multicolumn{2}{c}{\multirow{2}{*}{\makecell{CommNet}}} & 
        \multicolumn{2}{l}{$p=0$} & \multicolumn{2}{l}{\bm{$41.75\pm0.00$}}& \multicolumn{2}{l}{$40.74\pm0.23$} \\
        \multicolumn{2}{c}{} & 
        \multicolumn{2}{l}{$p=-0.5$} & \multicolumn{2}{l}{\bm{$35.38\pm1.28$}} & \multicolumn{2}{l}{$31.91\pm0.94$} \\
        
        
        

        \cmidrule(lr){1-2}
        \cmidrule(lr){3-4}
        \cmidrule(lr){5-8}
        
        \multicolumn{2}{c}{\multirow{2}{*}{\makecell{TarMAC}}} & 
        \multicolumn{2}{l}{Easy} & \multicolumn{2}{l}{\bm{$-4.08\pm0.04$}} & \multicolumn{2}{l}{$-4.24\pm0.08$} \\
        \multicolumn{2}{c}{} & 
        \multicolumn{2}{l}{Hard} & \multicolumn{2}{l}{\bm{$-9.33\pm0.29$}} & \multicolumn{2}{l}{$-9.80\pm0.08$}\\
        
        \cmidrule(lr){1-2}
        \cmidrule(lr){3-4}
        \cmidrule(lr){5-8}
        
        \multicolumn{2}{c}{\multirow{2}{*}{\makecell{NDQ}}} &  
        \multicolumn{2}{l}{3bane\_vs\_hM} & \multicolumn{2}{l}{\bm{$12.36\pm0.26$}} & \multicolumn{2}{l}{$11.32\pm0.15$} \\
        \multicolumn{2}{c}{} & 
        \multicolumn{2}{l}{1o\_2r\_vs\_4r} & \multicolumn{2}{l}{\bm{$17.70\pm0.16$}}  & \multicolumn{2}{l}{$16.33\pm0.54$}\\

        \bottomrule
    \end{tabular}
    \caption{Expected utilities for the defender trained with $\mathfrak{R}$-MACRL and the vanilla approach.}\label{table:psro_br_utilities_one_attacker}
    \vspace{-0.7cm}
\end{table}

\noindent
\textbf{Improving robustness.} 
To illustrate the improvement in robustness, we make comparisons between the defender trained by $\mathfrak{R}$-MACRL and the vanilla defending method. For the defender trained with $\mathfrak{R}$-MACRL, a population of attack policies are learned together with the defender, whose policy is also a mixture of sub-policies. In the vanilla training method, only a single defending policy is trained for the defender. We use the expected utility value (the accumulated team return) as the metric to compare the performance of the defender. Specifically, the larger the expected utility, the better the robustness. We denote the expected utility of the defender trained by $\mathfrak{R}$-MACRL as $u^{\operatorname{\mathfrak{R}}}_{\zeta}$ and the result for the vanilla one as $u^{\operatorname{vn}}_{\zeta}$. As shown in Table \ref{table:psro_br_utilities_one_attacker}, $\mathfrak{R}$-MACRL consistently outperforms the vanilla method over all the algorithms and environments. The improvement of expected utilities indicates that the defender trained by $\mathfrak{R}$-MACRL is more robust. Intuitively, the defender benefits from exploring a wider range of the policy space with $\mathfrak{R}$-MACRL, which enables the defender to maintain multi-agent coordination when faced with attacks.

\subsection{Scaling to Multiple Attackers}\label{sec:attack_results_2}
\begin{figure}
\centering
  \includegraphics[width=0.95\linewidth]{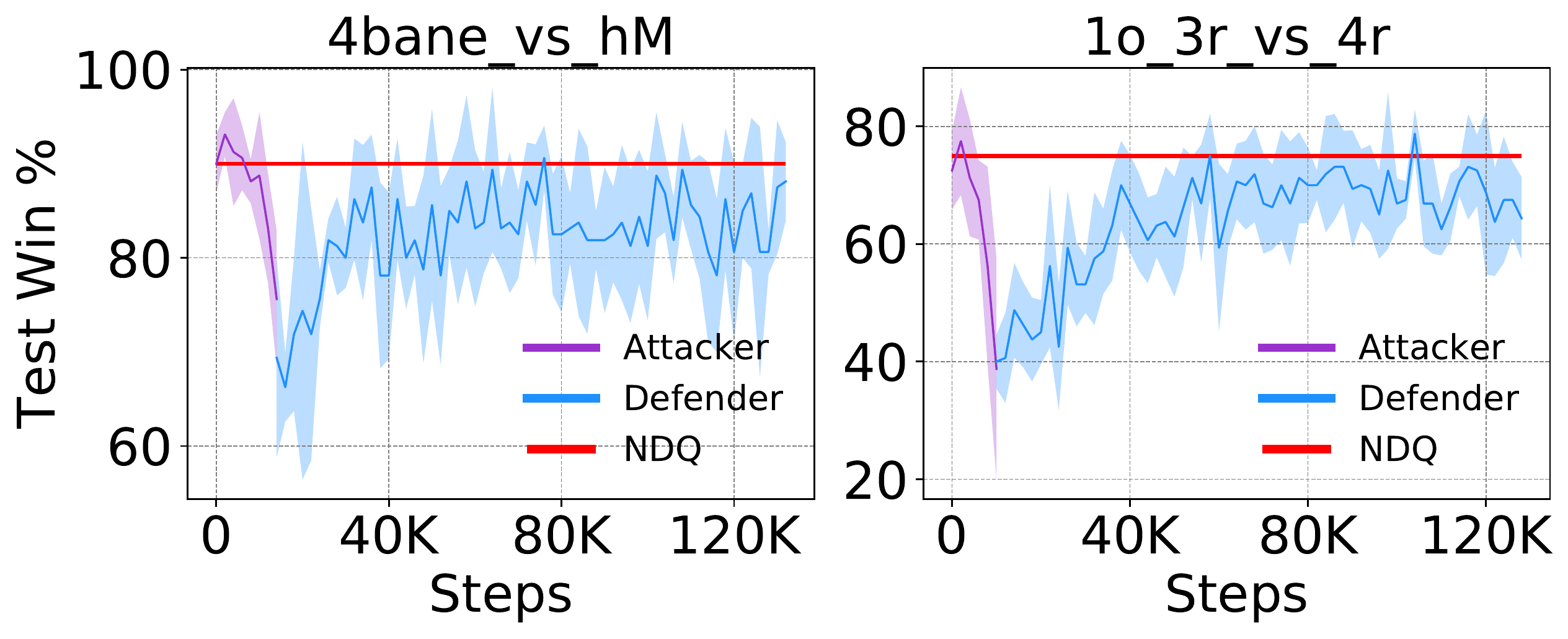} 
  \vspace{-0.3cm}
  \caption{\textit{Multiple attackers:} Attack and defence on NDQ.}
  \label{fig:ndq_results_2attackers}
\end{figure}

To examine the performance of the method in scenarios where there is more than one attacker, we conduct experiments on NDQ in two new challenge tasks: 4bane\_vs\_hM and $\operatorname{1o\_3r\_vs\_4r}$. In the former maps, i.e., 3bane\_vs\_hM and $\operatorname{1o\_2r\_vs\_4r}$, there are only three agents in the team, making multiple attackers unreasonable. To mitigate this, we create 4bane\_vs\_hM and $\operatorname{1o\_3r\_vs\_4r}$ based on 3bane\_vs\_hM and $\operatorname{1o\_2r\_vs\_4r}$ by increasing one more agent to the team. We randomly sample two agents to be malicious. According to our assumptions (agents have no knowledge about who are malicious and malicious agents cannot communicate with each other), we can directly apply the attacking method to each of the malicious agents. 
As shown in Fig. \ref{fig:ndq_results_2attackers}, the attackers can quickly learn effective adversarial policies with less learning steps. On the other hand, the message filter can still successfully recover multi-agent cooperation under multiple attackers, though it takes much more learning steps to converge. We further evaluate the effectiveness of $\mathfrak{R}$-MACRL on the two tasks and consistent results are observed: trained with $\mathfrak{R}$-MACRL, the agents can obtain larger expected utilities. More results on other algorithms and environments are provided in Appendix~\ref{appendix:more_results}.

\begin{table}
    \centering
    \begin{tabular}{crcrcrcr}
        \toprule
        \multicolumn{2}{c}{Methods} &
        \multicolumn{2}{l}{Scenarios} &
        \multicolumn{2}{l}{$u^{\operatorname{\mathfrak{R}}}_{\zeta}$} &
        \multicolumn{2}{l}{$u^{vn}_{\zeta}$}\\
        \cmidrule(lr){1-2}
        \cmidrule(lr){3-4}
        \cmidrule(lr){5-8}
        
        \multicolumn{2}{c}{\multirow{2}{*}{\makecell{NDQ}}} &  
        \multicolumn{2}{l}{4bane\_vs\_hM} & \multicolumn{2}{l}{\bm{$15.86\pm0.08$}}  & \multicolumn{2}{l}{$15.50\pm0.29$} \\
        \multicolumn{2}{c}{} & 
        \multicolumn{2}{l}{1o\_3r\_vs\_4r} & \multicolumn{2}{l}{\bm{$18.39\pm0.80$}}  & \multicolumn{2}{l}{$17.03\pm0.53$} \\


        
        
        

        

        \bottomrule
    \end{tabular}
    \caption{\textit{Multiple attackers:} Expected utilities for the defender trained with $\mathfrak{R}$-MACRL and the vanilla approach.}\label{table:psro_br_utilities_two_attackers}
    \vspace{-0.5cm}
\end{table}



\subsection{Ablation Study and Analysis}
We have shown that by integrating the message filter into the framework of $\mathfrak{R}$-MACRL, the robustness of MACRL algorithms can be improved. However, the performance of $\mathfrak{R}$-MACRL directly is affected by the message filter.
In this part, we will take a deeper look at how the message filter works.
\begin{figure*}[h]
    \centering
    \includegraphics[width=0.9\linewidth]{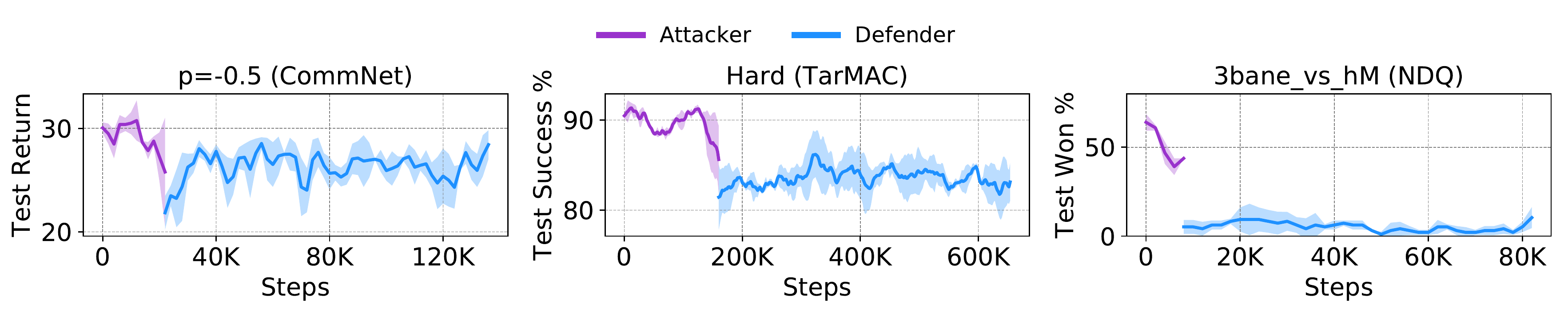}
    \vspace{-0.3cm}
    \caption{Defending by the message filter with the disabled message reconstructor.} \label{fig:psro_abl_1}
    \vspace{-0.3cm}
\end{figure*}
\begin{figure*}[h]
    \centering
    \includegraphics[width=0.9\linewidth]{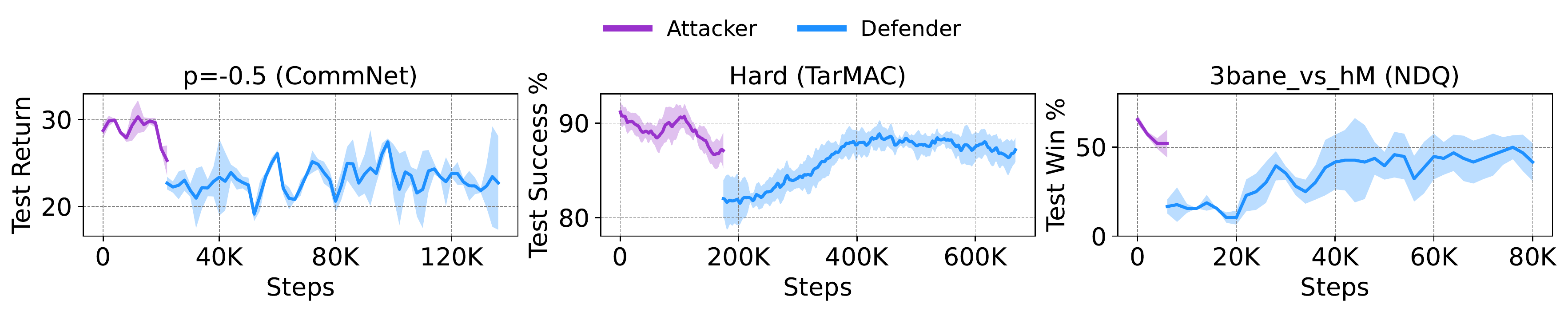}
    \vspace{-0.3cm}
    \caption{Defending by the message filter with the disabled anomaly detector.} \label{fig:psro_abl_2}
    \vspace{-0.cm}
\end{figure*}

\noindent
\textbf{Which components contribute to the performance?}
The message filter consists of two components: the anomaly detector and the message reconstructor. Here, we alternatively disable these two components in the message filter to study how they contribute to the performance. We run experiments on three scenarios, with each scenario corresponding to an MACRL algorithm. Following the former training procedure, we first train the original MACRL algorithm to obtain the action policy and the message policy; then we sample an attacker and train its adversarial policy by PPO; next we freeze the policy of the attacker and train the message filter (with the anomaly detector or the reconstructor disabled). As in Fig. \ref{fig:psro_abl_1}, if we disable the message reconstructor in the message filter and replace it with a random message generator, after being attacked, multi-agent coordination is hard to be recovered, demonstrating the importance of the reconstructor. We further disable the anomaly detector and randomly choose an agent to reconstruct its message, as in Fig. \ref{fig:psro_abl_2}, the defending performance is also poor. The ablation illustrates that both the anomaly detector and the reconstructor are critical in the message filter.

\begin{figure*}[ht]
    \centering
    \includegraphics[width=0.9\linewidth]{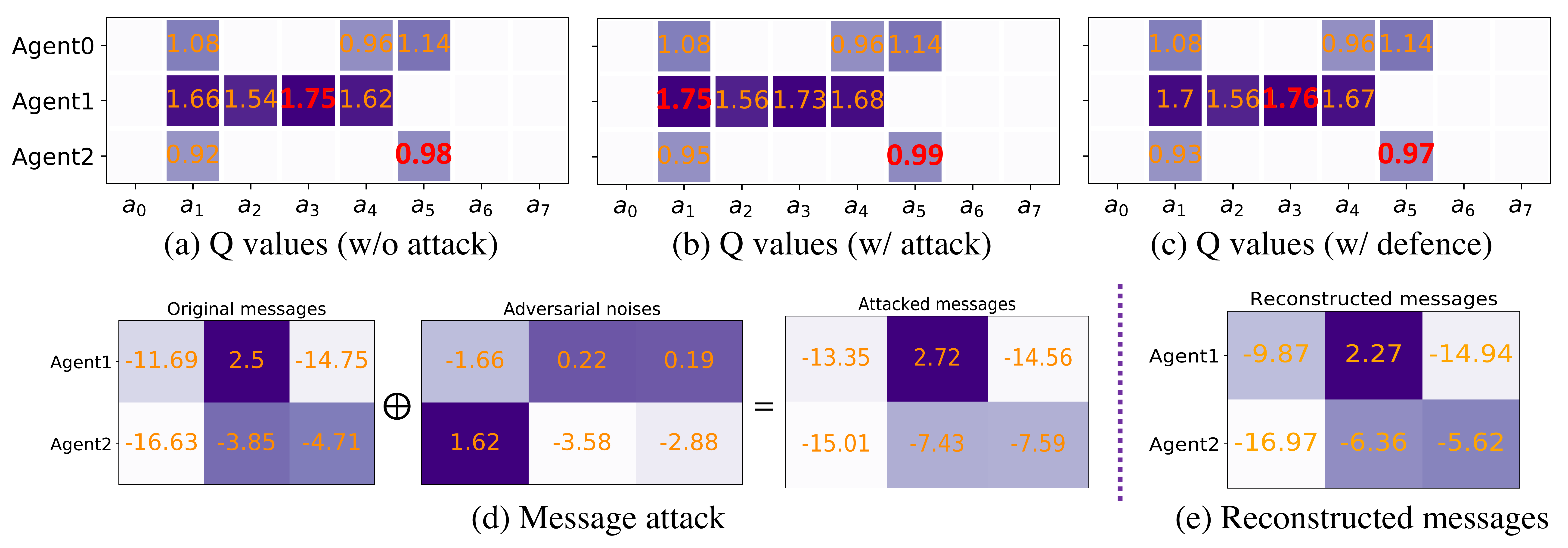}
    \caption{Q values and received messages under attack and defence. (a)-(c): Under attack, the optimal action of the benign agent $1$ shifts from $a_3$ to $a_1$. After applying the message filter, its optimal action is recovered. (d)-(e): The attacked messages become quite different from the original messages, while the reconstructed messages are to similar the original.} \label{fig:psro_analysis_1}
\end{figure*}

\noindent
\textbf{How does the proposed method work?}
Now we take a deeper look at how the message filter works. We conduct experiments on NDQ in the $\operatorname{3bane\_vs\_hM}$ task. 
  There are three agents in the team, of which agent $0$ is the attacker and the others, namely agent $1$ and agent $2$, are benign agents. As shown in Fig. \ref{fig:psro_analysis_1} (a)-(c), the action space of each agent contains eight elements. White cells in Q values correspond to illegal actions at the current state, and the action with red Q value is optimal, e.g., $a_5$ of agent $2$. When attacked by the agent $0$, the optimal action of agent $1$ shifts from $a_3$ to $a_1$, leading to sub-optimality. After applying the message filter, the decision of agent $1$ is corrected to $a_3$. We further examine the messages received by agent $1$ and agent $2$. As in Fig. \ref{fig:psro_analysis_1} (d)-(e), the attacked messages are quite different from the original messages, while after applying the message filter, the distance between the original messages and the reconstructed messages becomes closer.

\section{Conclusion}
In this paper, we systematically examine the problem of adversarial communication in MACRL. The problem is of importance but has been largely neglected before. We first provide the formulation of adversarial communication. Then we propose an effective method to model message attacks in MACRL. Following that, we design a two-stage message filter to defend against message attacks. Finally, to improve robustness, we formulate the adversarial communication problem as a two-player zero-sum game and propose $\mathfrak{R}$-MACRL to improve the robustness. Experiments on a wide range of algorithms and tasks show that many state-of-the-art MACRL methods are vulnerable to message attacks, while our algorithm can consistently recover multi-agent cooperation and improve the robustness of MACRL algorithms under message attacks.
\bibliographystyle{ACM-Reference-Format} 
\bibliography{ref}

\input{appendix}

\end{document}

%% file: appendix.tex
\clearpage
\newpage
\appendix

\section{Introduction to the selected MACRL algorithms}
\label{app:intro_to_algo}
\begin{figure*}
    \centering
    \includegraphics[width=0.85\linewidth]{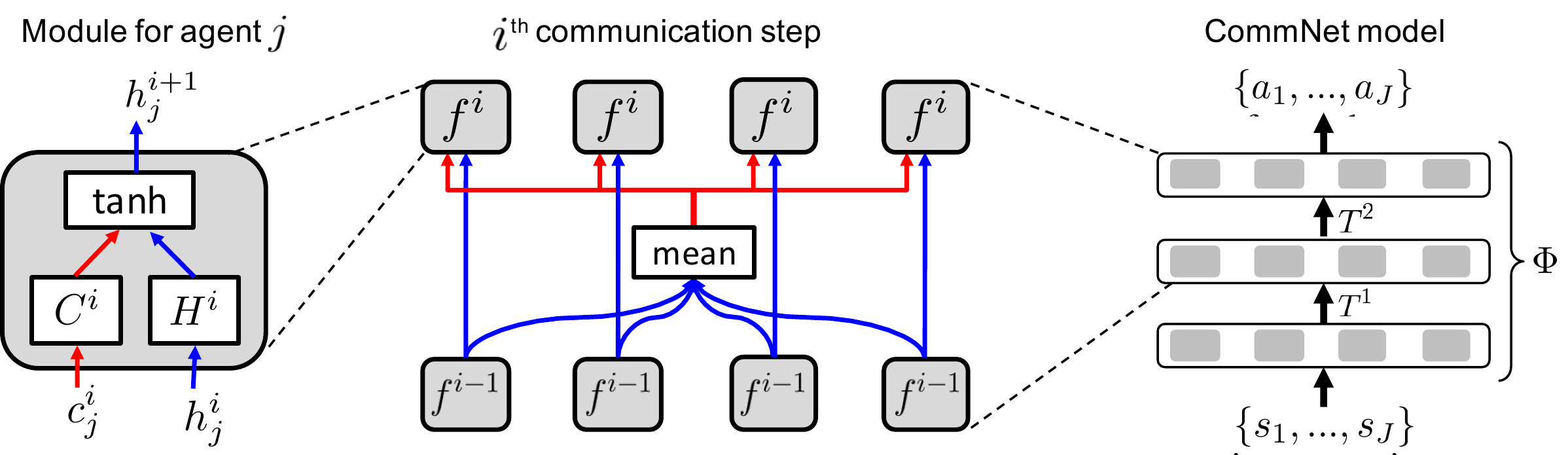}
    \caption{The Architecture of CommNet from~\cite{sukhbaatar2016learning}. An overview of our CommNet model. Left: view of module $f^i$ for a single agent $j$. Note that the parameters are shared across all agents. Middle: a single communication step, where each agents modules propagate their internal state $h$, as well as broadcasting a communication vector $c$ on a common channel (shown in red). Right: full model $\Phi$, showing input states $s$ for each agent, two communication steps and the output actions for each agent.} \label{fig:commnet_arch}
\end{figure*}

In this section, we introduce the selected MACRL algorithms for evaluation:

\noindent \textbf{CommNet~\cite{sukhbaatar2016learning}} is a simple yet effective multi-agent communicative algorithm where incoming messages of each agent are generated by averaging the messages sent by all agents in the last time step. We present the architecture of CommNet in Fig.~\ref{fig:commnet_arch}. 

\begin{figure*}
    \centering
    \includegraphics[width=0.85\linewidth]{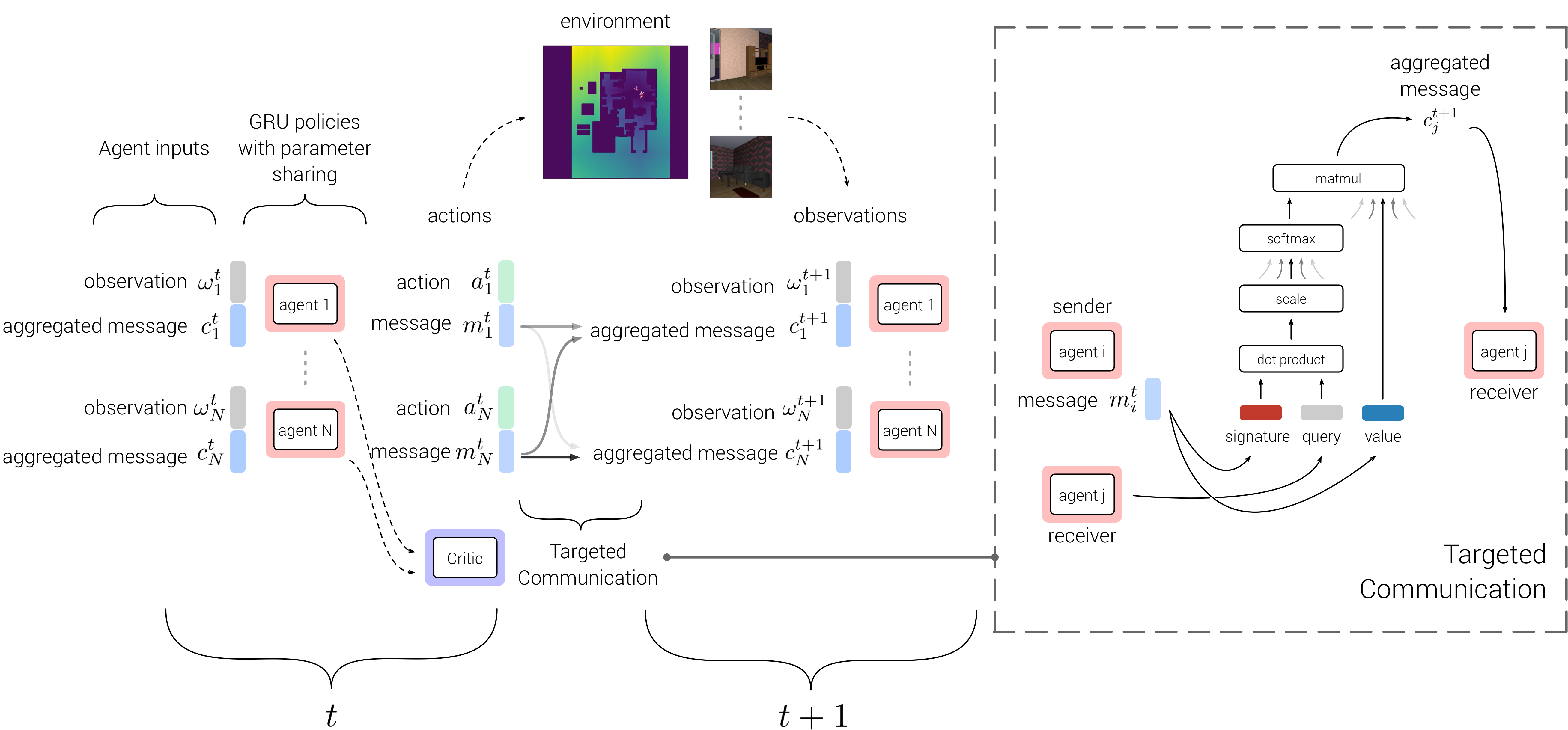}
    \caption{Overview of TarMAC from~\cite{das2019tarmac}.
    Left: At every timestep, each agent policy gets a local observation $\omega_i^t$
    and aggregated message $c_i^t$ as input, and predicts an environment action $a_i^t$
    and a targeted communication message $m_i^t$. Right: Targeted communication
    between agents is implemented as a signature-based soft attention mechanism.
    Each agent broadcasts a message $m_i^t$ consisting of a signature $k_i^t$,
    which can be used to encode agent-specific information and
    a value $v_i^t$, which contains the actual message. At the next timestep, each receiving agent
    gets as input a convex combination of message values, where the
    attention weights are obtained by a dot product between sender's signature $k_i^t$
    and a query vector $q_j^{t+1}$ predicted from the receiver's hidden state.} \label{fig:tarmac_arch}
\end{figure*}

\noindent \textbf{TarMAC~\cite{das2019tarmac}} extends CommNet by allowing agents to pay attention to important parts of the incoming messages via Attention~\cite{vaswani2017attention} network. Concretely, each agent generates signature and query vectors when sending message. In the message receiving phase, attention weights of incoming messages are calculated by comparing the similarity between the query vector and the signature vector of each incoming message. Then a weighted sum over all incoming messages is performed to determine the message an agent will receive for decentralized execution. The architecture of TarMAC is in Fig.~\ref{fig:tarmac_arch}.

\begin{figure*}
    \centering
    \includegraphics[width=0.85\linewidth]{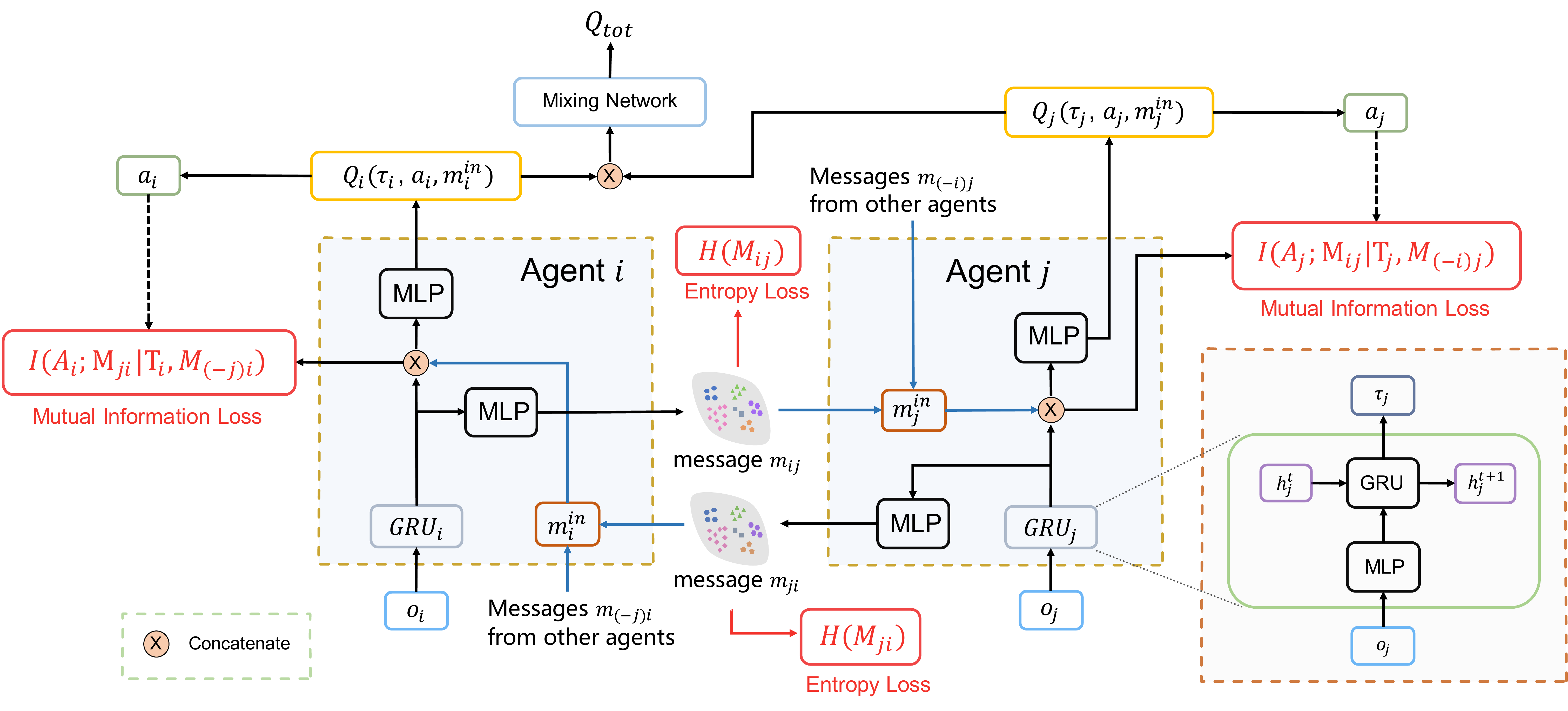}
    \caption{Overview of NDQ from~\cite{ndq2020iclr}. The message encoder generates an embedding distribution that is sampled and concatenated with the current local history to serve as an input to the local action-value function. Local action values are fed into a mixing network to to get an estimation of the global action value..} \label{fig:ndq_arch}
\end{figure*}

\noindent \textbf{NDQ~\cite{ndq2020iclr}} achieves nearly decomposable Q-functions via communication minimization. Specifically, it trains the communication model by maximizing mutual information between agents’ action selection and communication messages while minimizing the entropy of messages between agents. Each agent broadcasts messages to all other agents. The sent messages are sampled from learned distributions which are optimized by mutual information. The overview of NDQ is in Fig.~\ref{fig:ndq_arch}.



\section{The SMAC maps}
\label{app:scii_map}
\begin{figure}[ht]
\centering
\hfill
\includegraphics[width=0.23\textwidth]{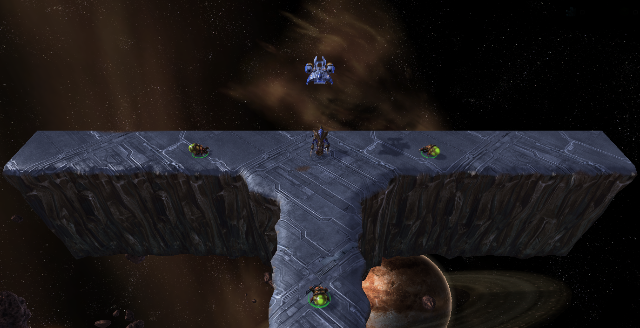}
\hfill
\centering
\includegraphics[width=0.23\textwidth]{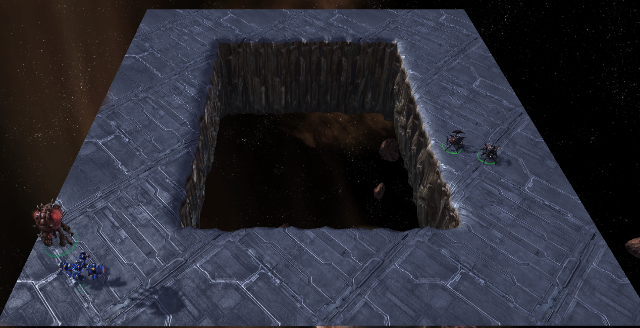}
\hfill
\centering
\caption{Snapshots of the StarCraft II scenarios. Left: 3bane\_vs\_hM. Right: 1o\_2r\_vs\_4r.}
\end{figure}
For the SCII environments, we conduct experiments on the following four maps to examine the attack and defence on NDQ: 

\textbf{3bane\_vs\_hM:} 3 Banelings try to kill a Hydralisk assisted by a Medivac. 3 Banelings together can just blow up the Hydralisk. Therefore, to win the game, 3 Banelings ally units should not give the Hydralisk changes to rest time when the Medivac can restore its health. 

\textbf{4bane\_vs\_hM:} Similar to 3bane\_vs\_hM, the main difference in 3bane\_vs\_hM is there are 4 Banelings.

\textbf{1o\_2r\_vs\_4r:} Ally units consist of 1 Overseer and 2 Roaches. The Overseer can find 4 Reapers and then notify its teammates to kill the invading enemy units, the Reapers, in order to win the game. At the start of an episode, ally units spawn at a random point on the map while enemy units are initialized at another random point. Given that only the Overseer knows the position of the enemy unit, the ability to learn to deliver this message to its teammates is vital for effectively winning the combat.

\textbf{1o\_3r\_vs\_4r:} Similar to 1o\_2r\_vs\_4r, the main difference in 1o\_3r\_vs\_4r is there are 1 Overseer and 3 Roaches.

\section{Hyperparameters}

We train all the selected MACRL methods with the same hyperparameters as in the corresponding papers, to reproduce the performances.
We use Adam~\cite{kingma2014adam} optimizer with a learning rate of 0.001 to train the attacker model as well as the anomaly detector and the message reconstructor. We use the default hyperparameters of Adam optimizer.  
$\alpha$, $\beta_1$ and $\beta_2$ are all 0.001. The clip ratio in PPO is $0.5$. The attacker model contains 3 linear layers and each layer has 64 neurons with ReLU activation. Same deep neural networks are used in the anomaly detector and the message reconstructor. We use PyTorch~\cite{pytorch19} to implement all the deep neural network models. 

\section{Additional Results}\label{appendix:more_results}
\begin{figure}[h]
    \centering
    \includegraphics[width=0.95\linewidth]{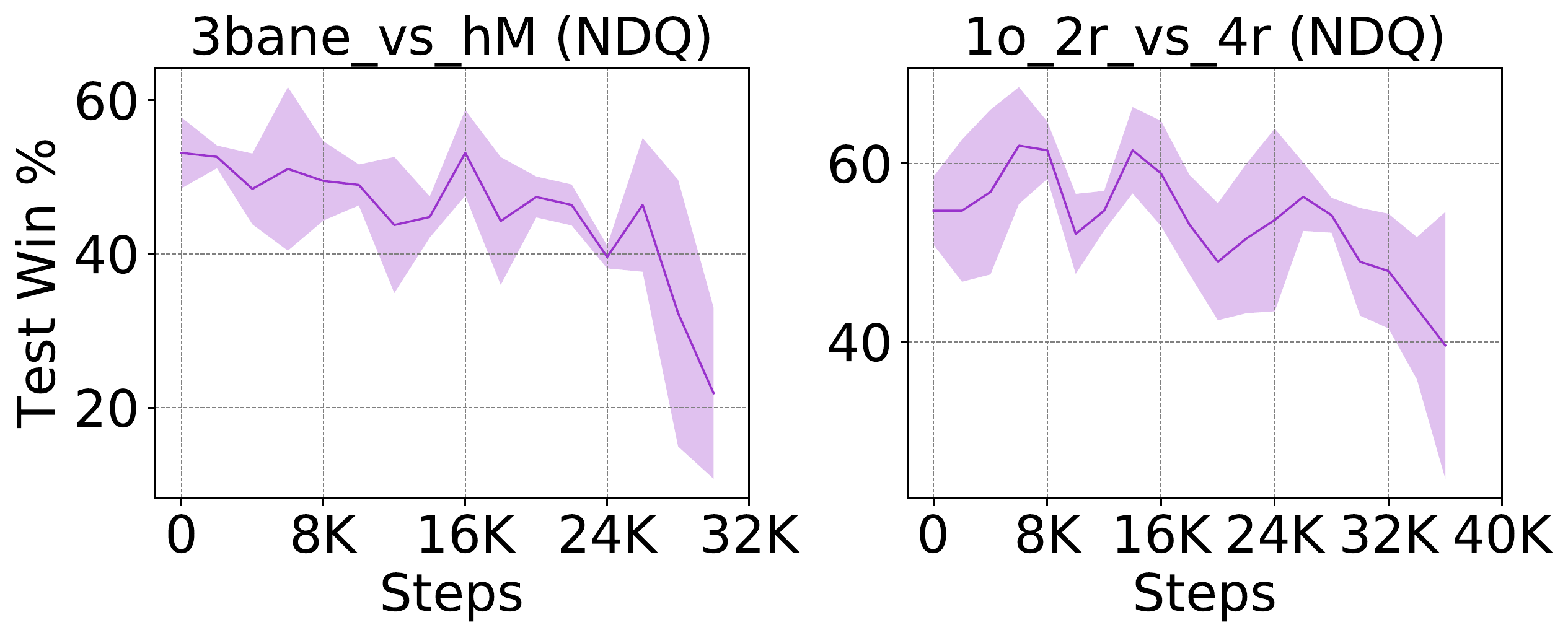}
    \caption{Exploiting the message filter in NDQ.}
    \label{app:exp_ndq}
\end{figure}

We train an adaptive attacker in NDQ to exploit the message filter. As in Fig. \ref{app:exp_ndq}, the test won rate decrease significantly, illustrating that a single message filter is brittle and can easily be exploited.

We also conduct experiments of multi attackers on the original environments. Two agents are selected randomly to be malicious. As in Table. \ref{table:appendix_psro_br_utilities_two_attackers_2}, $\mathfrak{R}$ outperforms the vanilla method in most of the tasks, demonstrating the effectiveness of our method. 
\begin{table}[ht]
    \centering
    \begin{tabular}{crcrcrcr}
        \toprule
        \multicolumn{2}{c}{Methods} &
        \multicolumn{2}{l}{Scenarios} &
        \multicolumn{2}{l}{$u^{\operatorname{\mathfrak{R}}}_{\zeta}$} &
        \multicolumn{2}{l}{$u^{vn}_{\zeta}$}\\

        \cmidrule(lr){1-2}
        \cmidrule(lr){3-4}
        \cmidrule(lr){5-8}
        
        \multicolumn{2}{c}{\multirow{2}{*}{\makecell{CommNet}}} & 
        \multicolumn{2}{l}{$p=0$} & \multicolumn{2}{l}{$41.64\pm1.90$}& \multicolumn{2}{l}{$41.25\pm1.21$} \\
        \multicolumn{2}{c}{} & 
        \multicolumn{2}{l}{$p=-0.5$} & \multicolumn{2}{l}{\bm{$36.83\pm1.71$}} & \multicolumn{2}{l}{$35.27\pm0.66$} \\
        
        
        

        \cmidrule(lr){1-2}
        \cmidrule(lr){3-4}
        \cmidrule(lr){5-8}
        
        \multicolumn{2}{c}{\multirow{2}{*}{\makecell{TarMAC}}} & 
        \multicolumn{2}{l}{Easy} & \multicolumn{2}{l}{\bm{$-0.70\pm0.00$}} & \multicolumn{2}{l}{$-0.80\pm0.00$} \\
        \multicolumn{2}{c}{} & 
        \multicolumn{2}{l}{Hard} & \multicolumn{2}{l}{\bm{$-0.90\pm0.00$}} & \multicolumn{2}{l}{$-1.00\pm0.00$}\\
        
         \cmidrule(lr){1-2}
        \cmidrule(lr){3-4}
        \cmidrule(lr){5-8}

        \multicolumn{2}{c}{\multirow{2}{*}{\makecell{NDQ}}} &  
        \multicolumn{2}{l}{3bane\_vs\_hM} & \multicolumn{2}{l}{\bm{$12.14\pm0.35$}}  & \multicolumn{2}{l}{$11.46\pm0.23$} \\ 
        \multicolumn{2}{c}{} & 
        \multicolumn{2}{l}{1o\_2r\_vs\_4r} & \multicolumn{2}{l}{\bm{$17.72\pm0.84$}}  & \multicolumn{2}{l}{$16.77\pm0.62$} \\
        \bottomrule
    \end{tabular}
    \caption{\textit{Multiple attackers:} Expected utilities for the defender trained with $\mathfrak{R}$-MACRL and the vanilla approach.}\label{table:appendix_psro_br_utilities_two_attackers_2}
\end{table}


%% file: main.bbl

\begin{thebibliography}{44}


\ifx \showCODEN    \undefined \def \showCODEN     #1{\unskip}     \fi
\ifx \showDOI      \undefined \def \showDOI       #1{#1}\fi
\ifx \showISBNx    \undefined \def \showISBNx     #1{\unskip}     \fi
\ifx \showISBNxiii \undefined \def \showISBNxiii  #1{\unskip}     \fi
\ifx \showISSN     \undefined \def \showISSN      #1{\unskip}     \fi
\ifx \showLCCN     \undefined \def \showLCCN      #1{\unskip}     \fi
\ifx \shownote     \undefined \def \shownote      #1{#1}          \fi
\ifx \showarticletitle \undefined \def \showarticletitle #1{#1}   \fi
\ifx \showURL      \undefined \def \showURL       {\relax}        \fi
\providecommand\bibfield[2]{#2}
\providecommand\bibinfo[2]{#2}
\providecommand\natexlab[1]{#1}
\providecommand\showeprint[2][]{arXiv:#2}

\bibitem[\protect\citeauthoryear{Ahilan and Dayan}{Ahilan and Dayan}{2021}]%
        {ahilan2020correcting}
\bibfield{author}{\bibinfo{person}{Sanjeevan Ahilan} {and}
  \bibinfo{person}{Peter Dayan}.} \bibinfo{year}{2021}\natexlab{}.
\newblock \showarticletitle{Correcting Experience Replay for Multi-Agent
  Communication}.
\newblock \bibinfo{journal}{\emph{ICLR}} (\bibinfo{year}{2021}).
\newblock


\bibitem[\protect\citeauthoryear{Barreno, Nelson, Joseph, and Tygar}{Barreno
  et~al\mbox{.}}{2010}]%
        {barreno2010security}
\bibfield{author}{\bibinfo{person}{Marco Barreno}, \bibinfo{person}{Blaine
  Nelson}, \bibinfo{person}{Anthony~D Joseph}, {and} \bibinfo{person}{J~Doug
  Tygar}.} \bibinfo{year}{2010}\natexlab{}.
\newblock \showarticletitle{The security of machine learning}.
\newblock \bibinfo{journal}{\emph{Machine Learning}} \bibinfo{volume}{81},
  \bibinfo{number}{2} (\bibinfo{year}{2010}), \bibinfo{pages}{121--148}.
\newblock


\bibitem[\protect\citeauthoryear{Blumenkamp and Prorok}{Blumenkamp and
  Prorok}{2020}]%
        {blumenkamp2020emergence}
\bibfield{author}{\bibinfo{person}{Jan Blumenkamp} {and}
  \bibinfo{person}{Amanda Prorok}.} \bibinfo{year}{2020}\natexlab{}.
\newblock \showarticletitle{The emergence of adversarial communication in
  multi-agent reinforcement learning}.
\newblock \bibinfo{journal}{\emph{CoRL}} (\bibinfo{year}{2020}).
\newblock


\bibitem[\protect\citeauthoryear{B{\"o}hmer, Kurin, and Whiteson}{B{\"o}hmer
  et~al\mbox{.}}{2020}]%
        {bohmer2020dcg}
\bibfield{author}{\bibinfo{person}{Wendelin B{\"o}hmer},
  \bibinfo{person}{Vitaly Kurin}, {and} \bibinfo{person}{Shimon Whiteson}.}
  \bibinfo{year}{2020}\natexlab{}.
\newblock \showarticletitle{Deep coordination graphs}. In
  \bibinfo{booktitle}{\emph{ICML}}. \bibinfo{pages}{980--991}.
\newblock


\bibitem[\protect\citeauthoryear{Carlini and Wagner}{Carlini and
  Wagner}{2017}]%
        {carlini2017towards}
\bibfield{author}{\bibinfo{person}{Nicholas Carlini} {and}
  \bibinfo{person}{David Wagner}.} \bibinfo{year}{2017}\natexlab{}.
\newblock \showarticletitle{Towards evaluating the robustness of neural
  networks}. In \bibinfo{booktitle}{\emph{2017 IEEE Symposium on Security and
  Privacy (SP)}}. \bibinfo{pages}{39--57}.
\newblock


\bibitem[\protect\citeauthoryear{Das, Gervet, Romoff, Batra, Parikh, Rabbat,
  and Pineau}{Das et~al\mbox{.}}{2019}]%
        {das2019tarmac}
\bibfield{author}{\bibinfo{person}{Abhishek Das},
  \bibinfo{person}{Th{\'e}ophile Gervet}, \bibinfo{person}{Joshua Romoff},
  \bibinfo{person}{Dhruv Batra}, \bibinfo{person}{Devi Parikh},
  \bibinfo{person}{Mike Rabbat}, {and} \bibinfo{person}{Joelle Pineau}.}
  \bibinfo{year}{2019}\natexlab{}.
\newblock \showarticletitle{Tarmac: Targeted multi-agent communication}. In
  \bibinfo{booktitle}{\emph{ICML}}. \bibinfo{pages}{1538--1546}.
\newblock


\bibitem[\protect\citeauthoryear{Foerster, Assael, de~Freitas, and
  Whiteson}{Foerster et~al\mbox{.}}{2016}]%
        {foerster2016learning}
\bibfield{author}{\bibinfo{person}{Jakob~N Foerster}, \bibinfo{person}{Yannis~M
  Assael}, \bibinfo{person}{Nando de Freitas}, {and} \bibinfo{person}{Shimon
  Whiteson}.} \bibinfo{year}{2016}\natexlab{}.
\newblock \showarticletitle{Learning to communicate with Deep multi-agent
  reinforcement learning}. In \bibinfo{booktitle}{\emph{NeurIPS}}.
  \bibinfo{pages}{2145--2153}.
\newblock


\bibitem[\protect\citeauthoryear{Gleave, Dennis, Wild, Kant, Levine, and
  Russell}{Gleave et~al\mbox{.}}{2020}]%
        {Gleave2020Adversarial}
\bibfield{author}{\bibinfo{person}{Adam Gleave}, \bibinfo{person}{Michael
  Dennis}, \bibinfo{person}{Cody Wild}, \bibinfo{person}{Neel Kant},
  \bibinfo{person}{Sergey Levine}, {and} \bibinfo{person}{Stuart Russell}.}
  \bibinfo{year}{2020}\natexlab{}.
\newblock \showarticletitle{Adversarial Policies: Attacking Deep Reinforcement
  Learning}. In \bibinfo{booktitle}{\emph{ICLR}}.
\newblock


\bibitem[\protect\citeauthoryear{Goodfellow, Papernot, Huang, Duan, Abbeel, and
  Clark}{Goodfellow et~al\mbox{.}}{2017}]%
        {goodfellow2017attacking}
\bibfield{author}{\bibinfo{person}{Ian Goodfellow}, \bibinfo{person}{Nicolas
  Papernot}, \bibinfo{person}{Sandy Huang}, \bibinfo{person}{Yan Duan},
  \bibinfo{person}{Pieter Abbeel}, {and} \bibinfo{person}{Jack Clark}.}
  \bibinfo{year}{2017}\natexlab{}.
\newblock \showarticletitle{Attacking machine learning with adversarial
  examples}.
\newblock \bibinfo{journal}{\emph{OpenAI. https://blog. openai.
  com/adversarial-example-research}} (\bibinfo{year}{2017}).
\newblock


\bibitem[\protect\citeauthoryear{Huang, Joseph, Nelson, Rubinstein, and
  Tygar}{Huang et~al\mbox{.}}{2011}]%
        {huang2011adversarial}
\bibfield{author}{\bibinfo{person}{Ling Huang}, \bibinfo{person}{Anthony~D
  Joseph}, \bibinfo{person}{Blaine Nelson}, \bibinfo{person}{Benjamin~IP
  Rubinstein}, {and} \bibinfo{person}{J~Doug Tygar}.}
  \bibinfo{year}{2011}\natexlab{}.
\newblock \showarticletitle{Adversarial machine learning}. In
  \bibinfo{booktitle}{\emph{Proceedings of the 4th ACM Workshop on Security and
  Artificial Intelligence}}. \bibinfo{pages}{43--58}.
\newblock


\bibitem[\protect\citeauthoryear{H{\"u}ttenrauch, {\v{S}}o{\v{s}}i{\'c}, and
  Neumann}{H{\"u}ttenrauch et~al\mbox{.}}{2017}]%
        {huttenrauch2017guided}
\bibfield{author}{\bibinfo{person}{Maximilian H{\"u}ttenrauch},
  \bibinfo{person}{Adrian {\v{S}}o{\v{s}}i{\'c}}, {and}
  \bibinfo{person}{Gerhard Neumann}.} \bibinfo{year}{2017}\natexlab{}.
\newblock \showarticletitle{Guided deep reinforcement learning for swarm
  systems}.
\newblock \bibinfo{journal}{\emph{arXiv preprint arXiv:1709.06011}}
  (\bibinfo{year}{2017}).
\newblock


\bibitem[\protect\citeauthoryear{Jaques, Lazaridou, Hughes, Gulcehre, Ortega,
  Strouse, Leibo, and De~Freitas}{Jaques et~al\mbox{.}}{2019}]%
        {jaques2019social}
\bibfield{author}{\bibinfo{person}{Natasha Jaques}, \bibinfo{person}{Angeliki
  Lazaridou}, \bibinfo{person}{Edward Hughes}, \bibinfo{person}{Caglar
  Gulcehre}, \bibinfo{person}{Pedro Ortega}, \bibinfo{person}{DJ Strouse},
  \bibinfo{person}{Joel~Z Leibo}, {and} \bibinfo{person}{Nando De~Freitas}.}
  \bibinfo{year}{2019}\natexlab{}.
\newblock \showarticletitle{Social influence as intrinsic motivation for
  multi-agent deep reinforcement learning}. In
  \bibinfo{booktitle}{\emph{ICML}}. \bibinfo{pages}{3040--3049}.
\newblock


\bibitem[\protect\citeauthoryear{Jiang and Lu}{Jiang and Lu}{2018}]%
        {jiang2018learning}
\bibfield{author}{\bibinfo{person}{Jiechuan Jiang} {and}
  \bibinfo{person}{Zongqing Lu}.} \bibinfo{year}{2018}\natexlab{}.
\newblock \showarticletitle{Learning attentional communication for multi-agent
  cooperation}. In \bibinfo{booktitle}{\emph{NeurIPS}}.
  \bibinfo{pages}{7254--7264}.
\newblock


\bibitem[\protect\citeauthoryear{Kim, Moon, Hostallero, Kang, Lee, Son, and
  Yi}{Kim et~al\mbox{.}}{2019}]%
        {kim2019learning}
\bibfield{author}{\bibinfo{person}{Daewoo Kim}, \bibinfo{person}{Sangwoo Moon},
  \bibinfo{person}{David Hostallero}, \bibinfo{person}{Wan~Ju Kang},
  \bibinfo{person}{Taeyoung Lee}, \bibinfo{person}{Kyunghwan Son}, {and}
  \bibinfo{person}{Yung Yi}.} \bibinfo{year}{2019}\natexlab{}.
\newblock \showarticletitle{Learning to schedule communication in multi-agent
  reinforcement learning}.
\newblock \bibinfo{journal}{\emph{ICLR}} (\bibinfo{year}{2019}).
\newblock


\bibitem[\protect\citeauthoryear{Kim, Park, and Sung}{Kim
  et~al\mbox{.}}{2021}]%
        {kimcommunication}
\bibfield{author}{\bibinfo{person}{Woojun Kim}, \bibinfo{person}{Jongeui Park},
  {and} \bibinfo{person}{Youngchul Sung}.} \bibinfo{year}{2021}\natexlab{}.
\newblock \showarticletitle{Communication in multi-Agent reinforcement
  learning: Intention Sharing}.
\newblock \bibinfo{journal}{\emph{ICLR}} (\bibinfo{year}{2021}).
\newblock


\bibitem[\protect\citeauthoryear{Kingma and Ba}{Kingma and Ba}{2014}]%
        {kingma2014adam}
\bibfield{author}{\bibinfo{person}{Diederik~P Kingma} {and}
  \bibinfo{person}{Jimmy Ba}.} \bibinfo{year}{2014}\natexlab{}.
\newblock \showarticletitle{Adam: A method for stochastic optimization}.
\newblock \bibinfo{journal}{\emph{arXiv preprint arXiv:1412.6980}}
  (\bibinfo{year}{2014}).
\newblock


\bibitem[\protect\citeauthoryear{Kirrmann}{Kirrmann}{2015}]%
        {bf}
\bibfield{author}{\bibinfo{person}{Hubert Kirrmann}.}
  \bibinfo{year}{2015}\natexlab{}.
\newblock \bibinfo{booktitle}{\emph{Fault Tolerant Computing in Industrial
  Automation}}.
\newblock \bibinfo{publisher}{Switzerland: ABB Research Center}.
\newblock


\bibitem[\protect\citeauthoryear{Kraemer and Banerjee}{Kraemer and
  Banerjee}{2016}]%
        {kraemer2016multi}
\bibfield{author}{\bibinfo{person}{Landon Kraemer} {and}
  \bibinfo{person}{Bikramjit Banerjee}.} \bibinfo{year}{2016}\natexlab{}.
\newblock \showarticletitle{Multi-agent reinforcement learning as a rehearsal
  for decentralized planning}.
\newblock \bibinfo{journal}{\emph{Neurocomputing}}  \bibinfo{volume}{190}
  (\bibinfo{year}{2016}), \bibinfo{pages}{82--94}.
\newblock


\bibitem[\protect\citeauthoryear{Lanctot, Zambaldi, Gruslys, Lazaridou, Tuyls,
  P{\'e}rolat, Silver, and Graepel}{Lanctot et~al\mbox{.}}{2017}]%
        {lanctot2017unified}
\bibfield{author}{\bibinfo{person}{Marc Lanctot}, \bibinfo{person}{Vinicius
  Zambaldi}, \bibinfo{person}{Audr{\=u}nas Gruslys}, \bibinfo{person}{Angeliki
  Lazaridou}, \bibinfo{person}{Karl Tuyls}, \bibinfo{person}{Julien
  P{\'e}rolat}, \bibinfo{person}{David Silver}, {and} \bibinfo{person}{Thore
  Graepel}.} \bibinfo{year}{2017}\natexlab{}.
\newblock \showarticletitle{A unified game-theoretic approach to multiagent
  reinforcement learning}. In \bibinfo{booktitle}{\emph{NeurIPS}}.
  \bibinfo{pages}{4193–4206}.
\newblock


\bibitem[\protect\citeauthoryear{Lin, Hong, Liao, Shih, Liu, and Sun}{Lin
  et~al\mbox{.}}{2017}]%
        {lin2017tactics}
\bibfield{author}{\bibinfo{person}{Yen-Chen Lin}, \bibinfo{person}{Zhang-Wei
  Hong}, \bibinfo{person}{Yuan-Hong Liao}, \bibinfo{person}{Meng-Li Shih},
  \bibinfo{person}{Ming-Yu Liu}, {and} \bibinfo{person}{Min Sun}.}
  \bibinfo{year}{2017}\natexlab{}.
\newblock \showarticletitle{Tactics of adversarial attack on deep reinforcement
  learning agents}. In \bibinfo{booktitle}{\emph{IJCAI}}.
  \bibinfo{pages}{3756--3762}.
\newblock


\bibitem[\protect\citeauthoryear{McMahan, Gordon, and Blum}{McMahan
  et~al\mbox{.}}{2003}]%
        {mcmahan2003planning}
\bibfield{author}{\bibinfo{person}{H~Brendan McMahan},
  \bibinfo{person}{Geoffrey~J Gordon}, {and} \bibinfo{person}{Avrim Blum}.}
  \bibinfo{year}{2003}\natexlab{}.
\newblock \showarticletitle{Planning in the presence of cost functions
  controlled by an adversary}. In \bibinfo{booktitle}{\emph{ICML}}.
  \bibinfo{pages}{536--543}.
\newblock


\bibitem[\protect\citeauthoryear{Mitchell, Blumenkamp, and Prorok}{Mitchell
  et~al\mbox{.}}{2020}]%
        {mitchell2020gaussian}
\bibfield{author}{\bibinfo{person}{Rupert Mitchell}, \bibinfo{person}{Jan
  Blumenkamp}, {and} \bibinfo{person}{Amanda Prorok}.}
  \bibinfo{year}{2020}\natexlab{}.
\newblock \showarticletitle{Gaussian Process Based Message Filtering for Robust
  Multi-Agent Cooperation in the Presence of Adversarial Communication}.
\newblock \bibinfo{journal}{\emph{arXiv preprint arXiv:2012.00508}}
  (\bibinfo{year}{2020}).
\newblock


\bibitem[\protect\citeauthoryear{Muller, Omidshafiei, Rowland, Tuyls, Perolat,
  Liu, Hennes, Marris, Lanctot, Hughes, et~al\mbox{.}}{Muller
  et~al\mbox{.}}{2019}]%
        {muller2019generalized}
\bibfield{author}{\bibinfo{person}{Paul Muller}, \bibinfo{person}{Shayegan
  Omidshafiei}, \bibinfo{person}{Mark Rowland}, \bibinfo{person}{Karl Tuyls},
  \bibinfo{person}{Julien Perolat}, \bibinfo{person}{Siqi Liu},
  \bibinfo{person}{Daniel Hennes}, \bibinfo{person}{Luke Marris},
  \bibinfo{person}{Marc Lanctot}, \bibinfo{person}{Edward Hughes},
  {et~al\mbox{.}}} \bibinfo{year}{2019}\natexlab{}.
\newblock \showarticletitle{A generalized training approach for multiagent
  learning}. In \bibinfo{booktitle}{\emph{ICLR}}.
\newblock


\bibitem[\protect\citeauthoryear{Oliehoek, Amato, et~al\mbox{.}}{Oliehoek
  et~al\mbox{.}}{2016}]%
        {oliehoek2016concise}
\bibfield{author}{\bibinfo{person}{Frans~A Oliehoek},
  \bibinfo{person}{Christopher Amato}, {et~al\mbox{.}}}
  \bibinfo{year}{2016}\natexlab{}.
\newblock \bibinfo{booktitle}{\emph{A Concise Introduction to Decentralized
  {POMDPs}}}. Vol.~\bibinfo{volume}{1}.
\newblock \bibinfo{publisher}{Springer}.
\newblock


\bibitem[\protect\citeauthoryear{Oliehoek, Spaan, and Vlassis}{Oliehoek
  et~al\mbox{.}}{2008}]%
        {oliehoek2008optimal}
\bibfield{author}{\bibinfo{person}{Frans~A Oliehoek},
  \bibinfo{person}{Matthijs~TJ Spaan}, {and} \bibinfo{person}{Nikos Vlassis}.}
  \bibinfo{year}{2008}\natexlab{}.
\newblock \showarticletitle{Optimal and approximate Q-value functions for
  decentralized {POMDP}s}.
\newblock \bibinfo{journal}{\emph{JAIR}}  \bibinfo{volume}{32}
  (\bibinfo{year}{2008}), \bibinfo{pages}{289--353}.
\newblock


\bibitem[\protect\citeauthoryear{Papernot, McDaniel, Goodfellow, Jha, Celik,
  and Swami}{Papernot et~al\mbox{.}}{2017}]%
        {papernot2017practical}
\bibfield{author}{\bibinfo{person}{Nicolas Papernot}, \bibinfo{person}{Patrick
  McDaniel}, \bibinfo{person}{Ian Goodfellow}, \bibinfo{person}{Somesh Jha},
  \bibinfo{person}{Z~Berkay Celik}, {and} \bibinfo{person}{Ananthram Swami}.}
  \bibinfo{year}{2017}\natexlab{}.
\newblock \showarticletitle{Practical black-box attacks against machine
  learning}. In \bibinfo{booktitle}{\emph{Proceedings of the 2017 ACM on Asia
  Conference on Computer and Communications Security}}.
  \bibinfo{pages}{506--519}.
\newblock


\bibitem[\protect\citeauthoryear{Paszke, Gross, Massa, Lerer, Bradbury, Chanan,
  Killeen, Lin, Gimelshein, Antiga, Desmaison, Kopf, Yang, DeVito, Raison,
  Tejani, Chilamkurthy, Steiner, Fang, Bai, and Chintala}{Paszke
  et~al\mbox{.}}{2019}]%
        {pytorch19}
\bibfield{author}{\bibinfo{person}{Adam Paszke}, \bibinfo{person}{Sam Gross},
  \bibinfo{person}{Francisco Massa}, \bibinfo{person}{Adam Lerer},
  \bibinfo{person}{James Bradbury}, \bibinfo{person}{Gregory Chanan},
  \bibinfo{person}{Trevor Killeen}, \bibinfo{person}{Zeming Lin},
  \bibinfo{person}{Natalia Gimelshein}, \bibinfo{person}{Luca Antiga},
  \bibinfo{person}{Alban Desmaison}, \bibinfo{person}{Andreas Kopf},
  \bibinfo{person}{Edward Yang}, \bibinfo{person}{Zachary DeVito},
  \bibinfo{person}{Martin Raison}, \bibinfo{person}{Alykhan Tejani},
  \bibinfo{person}{Sasank Chilamkurthy}, \bibinfo{person}{Benoit Steiner},
  \bibinfo{person}{Lu Fang}, \bibinfo{person}{Junjie Bai}, {and}
  \bibinfo{person}{Soumith Chintala}.} \bibinfo{year}{2019}\natexlab{}.
\newblock \showarticletitle{PyTorch: An Imperative Style, High-Performance Deep
  Learning Library}. In \bibinfo{booktitle}{\emph{NeurIPS}}.
  \bibinfo{pages}{8026--8037}.
\newblock


\bibitem[\protect\citeauthoryear{Samvelyan, Rashid, de~Witt, Farquhar,
  Nardelli, Rudner, Hung, Torr, Foerster, and Whiteson}{Samvelyan
  et~al\mbox{.}}{2019}]%
        {samvelyan19smac}
\bibfield{author}{\bibinfo{person}{Mikayel Samvelyan}, \bibinfo{person}{Tabish
  Rashid}, \bibinfo{person}{Christian~Schroeder de Witt},
  \bibinfo{person}{Gregory Farquhar}, \bibinfo{person}{Nantas Nardelli},
  \bibinfo{person}{Tim G.~J. Rudner}, \bibinfo{person}{Chia-Man Hung},
  \bibinfo{person}{Philiph H.~S. Torr}, \bibinfo{person}{Jakob Foerster}, {and}
  \bibinfo{person}{Shimon Whiteson}.} \bibinfo{year}{2019}\natexlab{}.
\newblock \showarticletitle{{The} {StarCraft} {multi}-{agent} {challenge}}.
\newblock \bibinfo{journal}{\emph{CoRR}}  \bibinfo{volume}{abs/1902.04043}
  (\bibinfo{year}{2019}).
\newblock


\bibitem[\protect\citeauthoryear{Schulman, Wolski, Dhariwal, Radford, and
  Klimov}{Schulman et~al\mbox{.}}{2017}]%
        {schulman2017proximal}
\bibfield{author}{\bibinfo{person}{John Schulman}, \bibinfo{person}{Filip
  Wolski}, \bibinfo{person}{Prafulla Dhariwal}, \bibinfo{person}{Alec Radford},
  {and} \bibinfo{person}{Oleg Klimov}.} \bibinfo{year}{2017}\natexlab{}.
\newblock \showarticletitle{Proximal policy optimization algorithms}.
\newblock \bibinfo{journal}{\emph{arXiv preprint arXiv:1707.06347}}
  (\bibinfo{year}{2017}).
\newblock


\bibitem[\protect\citeauthoryear{Singh, Jain, and Sukhbaatar}{Singh
  et~al\mbox{.}}{2018}]%
        {singh2018learning}
\bibfield{author}{\bibinfo{person}{Amanpreet Singh}, \bibinfo{person}{Tushar
  Jain}, {and} \bibinfo{person}{Sainbayar Sukhbaatar}.}
  \bibinfo{year}{2018}\natexlab{}.
\newblock \showarticletitle{Learning when to communicate at scale in multiagent
  cooperative and competitive tasks}.
\newblock \bibinfo{journal}{\emph{ICLR}} (\bibinfo{year}{2018}).
\newblock


\bibitem[\protect\citeauthoryear{Singh, Kumar, and Lau}{Singh
  et~al\mbox{.}}{2020}]%
        {singh2020hierarchical}
\bibfield{author}{\bibinfo{person}{Arambam~James Singh},
  \bibinfo{person}{Akshat Kumar}, {and} \bibinfo{person}{Hoong~Chuin Lau}.}
  \bibinfo{year}{2020}\natexlab{}.
\newblock \showarticletitle{Hierarchical Multiagent Reinforcement Learning for
  Maritime Traffic Management}. In \bibinfo{booktitle}{\emph{AAMAS}}.
  \bibinfo{pages}{1278--1286}.
\newblock


\bibitem[\protect\citeauthoryear{Sukhbaatar, Szlam, and Fergus}{Sukhbaatar
  et~al\mbox{.}}{2016}]%
        {sukhbaatar2016learning}
\bibfield{author}{\bibinfo{person}{Sainbayar Sukhbaatar},
  \bibinfo{person}{Arthur Szlam}, {and} \bibinfo{person}{Rob Fergus}.}
  \bibinfo{year}{2016}\natexlab{}.
\newblock \showarticletitle{Learning multiagent communication with
  backpropagation}. In \bibinfo{booktitle}{\emph{NeurIPS}}.
  \bibinfo{pages}{2252--2260}.
\newblock


\bibitem[\protect\citeauthoryear{Szegedy, Zaremba, Sutskever, Bruna, Erhan,
  Goodfellow, and Fergus}{Szegedy et~al\mbox{.}}{2014}]%
        {szegedy2014intriguing}
\bibfield{author}{\bibinfo{person}{Christian Szegedy},
  \bibinfo{person}{Wojciech Zaremba}, \bibinfo{person}{Ilya Sutskever},
  \bibinfo{person}{Joan Bruna}, \bibinfo{person}{Dumitru Erhan},
  \bibinfo{person}{Ian Goodfellow}, {and} \bibinfo{person}{Rob Fergus}.}
  \bibinfo{year}{2014}\natexlab{}.
\newblock \showarticletitle{Intriguing properties of neural networks}. In
  \bibinfo{booktitle}{\emph{ICLR}}.
\newblock


\bibitem[\protect\citeauthoryear{Tencent}{Tencent}{2019}]%
        {tesla}
\bibfield{author}{\bibinfo{person}{Keen Security~Lab Tencent}.}
  \bibinfo{year}{2019}\natexlab{}.
\newblock \showarticletitle{Experimental Security Research of Tesla Autopilot}.
\newblock  (\bibinfo{year}{2019}).
\newblock
\urldef\tempurl%
\url{https://keenlab.tencent.com/en/whitepapers/Experimental_Security_Research_of_Tesla_Autopilot.pdf}
\showURL{%
\tempurl}


\bibitem[\protect\citeauthoryear{Tramèr, Kurakin, Papernot, Goodfellow, Boneh,
  and McDaniel}{Tramèr et~al\mbox{.}}{2018}]%
        {tramer2018ensemble}
\bibfield{author}{\bibinfo{person}{Florian Tramèr}, \bibinfo{person}{Alexey
  Kurakin}, \bibinfo{person}{Nicolas Papernot}, \bibinfo{person}{Ian
  Goodfellow}, \bibinfo{person}{Dan Boneh}, {and} \bibinfo{person}{Patrick
  McDaniel}.} \bibinfo{year}{2018}\natexlab{}.
\newblock \showarticletitle{Ensemble adversarial training: attacks and
  defenses}. In \bibinfo{booktitle}{\emph{ICLR}}.
\newblock


\bibitem[\protect\citeauthoryear{Tu, Wang, Wang, Manivasagam, Ren, and
  Urtasun}{Tu et~al\mbox{.}}{2021}]%
        {tu2021adversarial}
\bibfield{author}{\bibinfo{person}{James Tu}, \bibinfo{person}{Tsunhsuan Wang},
  \bibinfo{person}{Jingkang Wang}, \bibinfo{person}{Sivabalan Manivasagam},
  \bibinfo{person}{Mengye Ren}, {and} \bibinfo{person}{Raquel Urtasun}.}
  \bibinfo{year}{2021}\natexlab{}.
\newblock \showarticletitle{Adversarial attacks on multi-agent communication}.
\newblock \bibinfo{journal}{\emph{arXiv preprint arXiv:2101.06560}}
  (\bibinfo{year}{2021}).
\newblock


\bibitem[\protect\citeauthoryear{Vaswani, Shazeer, Parmar, Uszkoreit, Jones,
  Gomez, Kaiser, and Polosukhin}{Vaswani et~al\mbox{.}}{2017}]%
        {vaswani2017attention}
\bibfield{author}{\bibinfo{person}{Ashish Vaswani}, \bibinfo{person}{Noam
  Shazeer}, \bibinfo{person}{Niki Parmar}, \bibinfo{person}{Jakob Uszkoreit},
  \bibinfo{person}{Llion Jones}, \bibinfo{person}{Aidan~N Gomez},
  \bibinfo{person}{{\L}ukasz Kaiser}, {and} \bibinfo{person}{Illia
  Polosukhin}.} \bibinfo{year}{2017}\natexlab{}.
\newblock \showarticletitle{Attention is all you need}. In
  \bibinfo{booktitle}{\emph{NeurIPS}}. \bibinfo{pages}{5998--6008}.
\newblock


\bibitem[\protect\citeauthoryear{Vinyals, Babuschkin, Czarnecki, Mathieu,
  Dudzik, Chung, Choi, Powell, Ewalds, Georgiev, et~al\mbox{.}}{Vinyals
  et~al\mbox{.}}{2019}]%
        {vinyals2019grandmaster}
\bibfield{author}{\bibinfo{person}{Oriol Vinyals}, \bibinfo{person}{Igor
  Babuschkin}, \bibinfo{person}{Wojciech~M Czarnecki},
  \bibinfo{person}{Micha{\"e}l Mathieu}, \bibinfo{person}{Andrew Dudzik},
  \bibinfo{person}{Junyoung Chung}, \bibinfo{person}{David~H Choi},
  \bibinfo{person}{Richard Powell}, \bibinfo{person}{Timo Ewalds},
  \bibinfo{person}{Petko Georgiev}, {et~al\mbox{.}}}
  \bibinfo{year}{2019}\natexlab{}.
\newblock \showarticletitle{Grandmaster level in {StarCraft II} using
  multi-agent reinforcement learning}.
\newblock \bibinfo{journal}{\emph{Nature}} \bibinfo{volume}{575},
  \bibinfo{number}{7782} (\bibinfo{year}{2019}), \bibinfo{pages}{350--354}.
\newblock


\bibitem[\protect\citeauthoryear{Wang, He, Yu, Qiu, An, and Rabinovich}{Wang
  et~al\mbox{.}}{2020}]%
        {wang2020imac}
\bibfield{author}{\bibinfo{person}{Rundong Wang}, \bibinfo{person}{Xu He},
  \bibinfo{person}{Runsheng Yu}, \bibinfo{person}{Wei Qiu}, \bibinfo{person}{Bo
  An}, {and} \bibinfo{person}{Zinovi Rabinovich}.}
  \bibinfo{year}{2020}\natexlab{}.
\newblock \showarticletitle{Learning efficient multi-agent communication: An
  information bottleneck approach}. In \bibinfo{booktitle}{\emph{ICML}}.
  \bibinfo{pages}{9908--9918}.
\newblock


\bibitem[\protect\citeauthoryear{Wang, Wang, Zheng, and Zhang}{Wang
  et~al\mbox{.}}{2019}]%
        {ndq2020iclr}
\bibfield{author}{\bibinfo{person}{Tonghan Wang}, \bibinfo{person}{Jianhao
  Wang}, \bibinfo{person}{Chongyi Zheng}, {and} \bibinfo{person}{Chongjie
  Zhang}.} \bibinfo{year}{2019}\natexlab{}.
\newblock \showarticletitle{Learning Nearly Decomposable Value Functions Via
  Communication Minimization}. In \bibinfo{booktitle}{\emph{ICLR}}.
\newblock


\bibitem[\protect\citeauthoryear{Xu, Wang, Raizman, and Rabinovich}{Xu
  et~al\mbox{.}}{2021}]%
        {xu2021transferable}
\bibfield{author}{\bibinfo{person}{Hang Xu}, \bibinfo{person}{Rundong Wang},
  \bibinfo{person}{Lev Raizman}, {and} \bibinfo{person}{Zinovi Rabinovich}.}
  \bibinfo{year}{2021}\natexlab{}.
\newblock \showarticletitle{Transferable environment poisoning: Training-time
  attack on reinforcement learning}. In \bibinfo{booktitle}{\emph{AAMAS}}.
  \bibinfo{pages}{1398--1406}.
\newblock


\bibitem[\protect\citeauthoryear{Zhang, Chen, Boning, and Hsieh}{Zhang
  et~al\mbox{.}}{2021}]%
        {zhang2021robust}
\bibfield{author}{\bibinfo{person}{Huan Zhang}, \bibinfo{person}{Hongge Chen},
  \bibinfo{person}{Duane~S Boning}, {and} \bibinfo{person}{Cho-Jui Hsieh}.}
  \bibinfo{year}{2021}\natexlab{}.
\newblock \showarticletitle{Robust reinforcement learning on state observations
  with learned optimal adversary}. In \bibinfo{booktitle}{\emph{ICLR}}.
\newblock


\bibitem[\protect\citeauthoryear{Zhang, Chen, Xiao, Li, Liu, Boning, and
  Hsieh}{Zhang et~al\mbox{.}}{2020a}]%
        {zhang2020robust}
\bibfield{author}{\bibinfo{person}{Huan Zhang}, \bibinfo{person}{Hongge Chen},
  \bibinfo{person}{Chaowei Xiao}, \bibinfo{person}{Bo Li},
  \bibinfo{person}{Mingyan Liu}, \bibinfo{person}{Duane Boning}, {and}
  \bibinfo{person}{Cho-Jui Hsieh}.} \bibinfo{year}{2020}\natexlab{a}.
\newblock \showarticletitle{Robust deep reinforcement learning against
  adversarial perturbations on state observations}. In
  \bibinfo{booktitle}{\emph{NeurIPS}}. \bibinfo{pages}{21024--21037}.
\newblock


\bibitem[\protect\citeauthoryear{Zhang, Lin, and Zhang}{Zhang
  et~al\mbox{.}}{2020b}]%
        {zhang2020tmc}
\bibfield{author}{\bibinfo{person}{Sai~Qian Zhang}, \bibinfo{person}{Jieyu
  Lin}, {and} \bibinfo{person}{Qi Zhang}.} \bibinfo{year}{2020}\natexlab{b}.
\newblock \showarticletitle{Succinct and robust multi-agent communication with
  temporal message control}.
\newblock \bibinfo{journal}{\emph{arXiv preprint arXiv:2010.14391}}
  (\bibinfo{year}{2020}), \bibinfo{pages}{17271--17282}.
\newblock


\end{thebibliography}
